\documentclass{article}

\usepackage{arxiv}

\usepackage[utf8]{inputenc}
\usepackage[T1]{fontenc}

\usepackage{latexsym}
\usepackage{graphicx}
\usepackage{multicol,multirow}
\usepackage{amsmath,amsfonts,amsthm,amssymb,dsfont}
\usepackage{mathrsfs}
\usepackage{rotating}
\usepackage{appendix}
\usepackage{ifpdf}
\usepackage{times}
\usepackage{newtxmath}
\usepackage{textcomp}
\usepackage{xcolor}
\usepackage{hyperref}
\usepackage{lipsum}
\usepackage{cleveref}
\usepackage{caption,subcaption}
\usepackage{booktabs,tabularx}
\usepackage{csquotes}
\usepackage{svg}

\usepackage[
backend=biber,
style=numeric,
natbib=true,
url=false,
eprint=false,
sorting=none,
]{biblatex}
\graphicspath{ {./figs/} }

\makeatletter
\renewcommand\hyper@natlinkbreak[2]{#1}
\makeatother

\date{}

\title{Super-Resolving Coarse-Resolution Weather Forecasts with Flow Matching}
\author{
      {Aymeric Delefosse}$^1$,$\;$ {Anastase Charantonis}$^1$,$\;$ {Dominique Béréziat}$^2$ \\
      $^1$Inria, ARCHES, Paris, France \\
      $^2$Sorbonne Université, CNRS, LIP6, Paris, France
}

\begin{document}
\maketitle

\begin{abstract}
      Machine learning-based weather forecasting models now surpass state-of-the-art numerical weather prediction systems, but training and operating these models at high spatial resolution remains computationally expensive. We present a modular framework that decouples forecasting from spatial resolution by applying learned generative super-resolution as a post-processing step to coarse-resolution forecast trajectories. We formulate super-resolution as a stochastic inverse problem, using a residual formulation to preserve large-scale structure while reconstructing unresolved variability. The model is trained with flow matching exclusively on reanalysis data and is applied to global medium-range forecasts. We evaluate (i) design consistency by re-coarsening super-resolved forecasts and comparing them to the original coarse trajectories, and (ii) high-resolution forecast quality using standard ensemble verification metrics and spectral diagnostics. Results show that super-resolution preserves large-scale structure and variance after re-coarsening, introduces physically consistent small-scale variability, and achieves competitive probabilistic forecast skill at 0.25° resolution relative to an operational ensemble baseline, while requiring only a modest additional training cost compared with end-to-end high-resolution forecasting.
\end{abstract}

\section{Introduction}
In recent years, machine learning (ML)-based weather forecasting has advanced significantly, with models now matching or surpassing traditional numerical weather prediction (NWP) systems in forecast skill and inference efficiency \citep{biAccurateMediumrangeGlobal2023,lamLearningSkillfulMediumrange2023, langAIFSCRPSEnsembleForecasting2026}. This progress is attributable to the development of increasingly powerful architectures and the availability of high-quality, multi-decadal reanalysis datasets, such as ERA5 \citep{hersbachERA5GlobalReanalysis2020}. This new generation of models enables rapid generation of large ensembles, but training remains computationally demanding: multi-terabyte, high-resolution datasets impose severe data throughput and memory constraints, which typically require access to large-scale computational resources \citep{bonevFourCastNet3Geometric2025}.

At the same time, models trained at coarser spatial resolutions, where data volume and memory requirements are far more manageable, demonstrate that competitive forecast skill can still be achieved \citep{verma2024climode,nguyenScalingTransformerNeural2024,couaironArchesWeatherArchesWeatherGenDeterministic2024}. This reflects the scale dependence of atmospheric predictability: for medium-range forecasts, skill is primarily constrained by large-scale atmospheric dynamics \citep{anthesPredictabilityMesoscaleAtmospheric1985,mooreFlowDependenceMediumRange2023}. However, this reduced-cost regime remains insufficient to resolve the spatial scales that govern regional weather. These are necessary to represent the organization and intensity of the phenomena that drive local impacts, including heavy precipitation, strong winds, and tropical cyclones.

This motivates exploring alternative strategies that decouple spatial resolution from model training cost. Inspired by advances in computer vision, a natural direction is to employ learned super-resolution (SR) methods \citep{dongImageSuperResolutionUsing2016,ledigPhotoRealisticSingleImage2017,sahariaImageSuperResolutionIterative2023}, also commonly referred to as downscaling in the climate sciences. Rather than retraining or fine-tuning a forecasting model at higher resolution, SR aims to learn a mapping from coarse to fine spatial scales. This approach offers complementary advantages in modularity and efficiency: the forecasting and the downscaling models can be developed and optimized independently. In this framework, forecasting operates at low resolution, while super-resolution recovers subgrid-scale variability.
So far, data-driven downscaling has primarily focused on regional climate applications \citep{wattGenerativeDiffusionbasedDownscaling2024,mardaniResidualCorrectiveDiffusion2025,hessFastScaleadaptiveUncertaintyaware2025}, typically operating on individual states and a limited set of target variables, without explicit consideration of forecast trajectories. Notable exceptions include SwinRDM \citep{chenSwinRDMIntegrateSwinRNN2023}, which combines a coarse-resolution recurrent forecast model with a diffusion-based upsampling step, demonstrating the viability of cascading forecasting and generative downscaling for high-resolution weather prediction. More recently, Climate in a Bottle \citep{brenowitzClimateBottleGenerative2025} integrates generative downscaling within a climate modeling pipeline to produce kilometer-scale forecasts.

In this work, we apply learned generative super-resolution to global medium-range forecast trajectories produced by coarse-resolution data-driven models. Using ArchesWeather and ArchesWeatherGen \citep{couaironArchesWeatherArchesWeatherGenDeterministic2024} as a representative forecasting system, we assess whether super-resolution can reconstruct physically meaningful fine-scale structure while remaining consistent with the underlying large-scale dynamics, thereby supporting its feasibility as a modular and computationally affordable pathway toward higher-resolution forecasting. The remainder of the paper is structured as follows. We first introduce the problem formulation and the proposed super-resolution framework. We then describe the generative model and its application to forecast trajectories, followed by the experimental setup, evaluation methodology, and results. Finally, we discuss the implications, limitations, and possible directions for future work.

\section{Methods}\label{sec:methods}
\subsection{Problem formulation}\label{sec:problem_formulation}
We consider the problem of reconstructing a high-resolution atmospheric trajectory consistent with a given coarse-resolution forecast. Let $\mathbf{x}_t$ denote the global weather state at time $t$. In our framework, $\mathbf{x}_t$ consists of six upper-air variables (temperature, geopotential, specific humidity, wind components $U$, $V$ and $W$) sampled on a latitude-longitude grid at 13 pressure levels, and four surface variables (2m temperature, mean sea-level pressure, and 10m wind components $U$ and $V$).

\paragraph{Coarse-resolution forecasts.}
Let $\mathbf{x}_t^{\mathrm{LR}}$ denote a low-resolution (LR) atmospheric state at time $t$.
A forecasting model $f_\theta$ is trained to predict the future state at a fixed lead time $\delta$,
\begin{equation}
      \hat{\mathbf{x}}_{t+\delta}^{\mathrm{LR}} = f_\theta\!\left(\mathbf{x}_t^{\mathrm{LR}}\right),
\end{equation}
where $\delta$ is set to $\delta = 24\,\mathrm{h}$ throughout this work.
Forecast trajectories of length $T$ are obtained by applying the model autoregressively,
\begin{equation}
      \hat{\mathbf{x}}_{t+k\delta}^{\mathrm{LR}} = f_\theta\!\left(\hat{\mathbf{x}}_{t+(k-1)\delta}^{\mathrm{LR}}\right),
      \; 1 \le k \le T.
\end{equation}
Our objective is to construct corresponding high-resolution (HR) forecasts $\left(\hat{\mathbf{x}}_{t+k\delta}^{\mathrm{HR}}\right)_{1 \le k \le T}$.

\paragraph{The coarse-graining operator.}
The relationship between resolutions is formalized through a known regridding operator $\mathcal{C}$. For any high-resolution state $\mathbf{x}_t^{\mathrm{HR}}$, the associated coarse representation is given by
$
      \mathbf{x}_t^{\mathrm{LR}} = \mathcal{C}\!\left(\mathbf{x}_t^{\mathrm{HR}}\right).
$
By construction, $\mathcal{C}$ removes information associated with unresolved spatial scales and therefore defines a lossy, non-invertible mapping. The super-resolution problem can thus be viewed as learning an approximate inverse operator $\mathcal{C}^{-1}_\phi$ such that
$
      \hat{\mathbf{x}}_t^{\mathrm{HR}} = \mathcal{C}^{-1}_\phi\!\left(\mathbf{x}_t^{\mathrm{LR}}\right)
$.
A design objective is that reapplying the coarse-graining operator to the reconstructed high-resolution fields approximately recovers the original coarse state,
$
      \mathcal{C}\!\left(\hat{\mathbf{x}}_t^{\mathrm{HR}}\right) \approx \mathbf{x}_t^{\mathrm{LR}}.
$
This reflects the intent to preserve the large-scale organization of the coarse state while enriching it with unresolved structure.

\paragraph{Residual formulation.}
To structure this inverse problem, we adopt a residual decomposition formally analogous to a Reynolds decomposition, in which a field is expressed as the sum of a resolved component and a fluctuating part \citep{reynoldsIVDynamicalTheory1895}. Let $\mathcal{I}$ denote a deterministic interpolation operator (e.g., bicubic) mapping coarse-resolution fields to the high-resolution grid, and define
${
      \uparrow\!\mathbf{x}_t^{\mathrm{LR}} = \mathcal{I}\!\left(\mathbf{x}_t^{\mathrm{LR}}\right).
      }$
By construction, this interpolated field preserves the large-scale structure of the coarse state but contains no genuine small-scale variability, i.e., ${\mathcal{C}\!\left(\mathcal{I}\left(\mathbf{x}_t^{\mathrm{LR}}\right)\right) \approx \mathbf{x}_t^{\mathrm{LR}}}$. The high-resolution state can then be written as
${
      \mathbf{x}_t^{\mathrm{HR}} = \;\uparrow\!\mathbf{x}_t^{\mathrm{LR}} + \mathbf{r}_t,
      }$
where $\mathbf{r}$ represents the unresolved high-frequency residual. We train a neural network $g_\phi$ to predict this residual,
${
                  \hat{\mathbf{r}}_t = g_\phi\left(\uparrow\!\mathbf{x}_t^{\mathrm{LR}}\right),
            }$
yielding the final reconstruction
\begin{equation}
      \hat{\mathbf{x}}_t^{\mathrm{HR}} = \;\uparrow\!\mathbf{x}_t^{\mathrm{LR}} + \hat{\mathbf{r}}_t.
\end{equation}
This formulation concentrates model capacity on reconstructing fine-scale structure while explicitly preserving the large-scale information already present in the coarse field. Similar residual strategies have proven effective in climate downscaling \citep{liGenerativeEmulationWeather2024,mardaniResidualCorrectiveDiffusion2025,lopez-gomezDynamicalgenerativeDownscalingClimate2025}.

\paragraph{Stochastic inverse modeling.}
From a physical standpoint, the inverse problem is inherently non-unique: a coarse forecast constrains the large-scale atmospheric flow but does not uniquely determine fine-scale structure. Subgrid variability admits many realizations compatible with the same resolved state. Accordingly, our objective is not to recover a single deterministic field, but to learn a stochastic inverse mapping by modeling the conditional distribution of unresolved residuals, $p(\mathbf{r}_t \mid \,\uparrow \mathbf{x}_t^{\mathrm{LR}})$.
Although we operate on forecast trajectories, we do not explicitly condition the inverse model on temporal history. At 24-hour lead times and 1.5° resolution, fine-scale organization is primarily constrained by the large-scale flow at that time, making additional temporal conditioning weakly informative while increasing model complexity.

\subsection{Super-resolution model}\label{sec:sr_model}

We adopt a conditional generative super-resolution approach inspired by SR3 \citep{sahariaImageSuperResolutionIterative2023}, adapted to the residual formulation and to global weather fields. Instead of the original SR3 U-Net backbone, we use a 3D Swin U-Net transformer architecture as in ArchesWeather \citep{couaironArchesWeatherEfficientAI2024}, which better scales to high-resolution fields \citep{crowsonScalableHighResolutionPixelSpace2024}. To model the stochasticity of unresolved scales, we train the model using flow matching \citep{lipmanFlowMatchingGenerative2023}, learning the conditional distribution
$p(\mathbf{r}_t \mid \, \uparrow\mathbf{x}_t^{\mathrm{LR}})$. Further architectural details are provided in Appendix~\ref{app:model_architecture}.

\begin{figure}[htb]
      \centering
      \includegraphics[width=\textwidth]{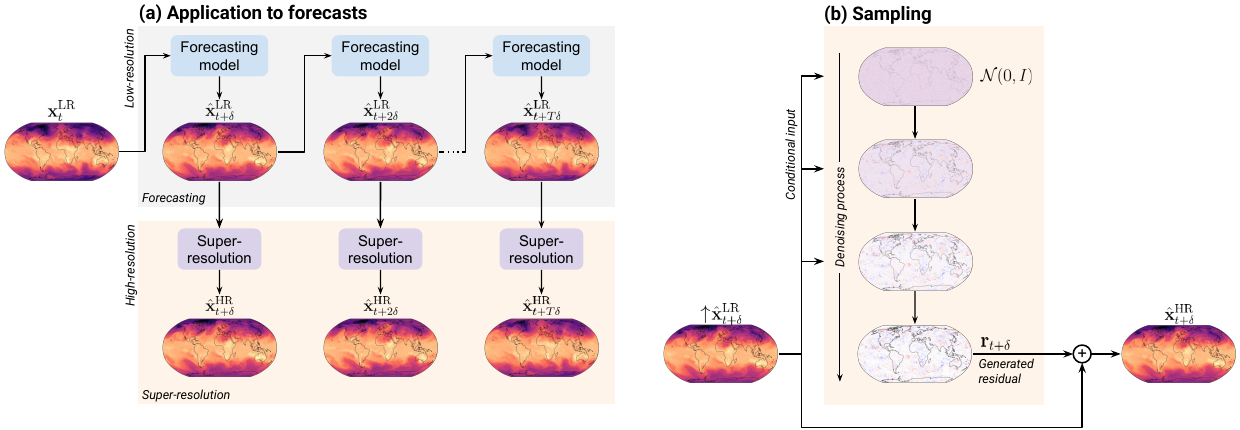}
      \caption{\textbf{Overview of the super-resolution procedure.} (a) Super-resolution is applied independently to each coarse-resolution forecast state produced by the base forecasting model. (b) Sampling procedure: starting from Gaussian noise, a flow matching model generates a high-resolution residual conditioned on the interpolated coarse-resolution forecast, which is added to obtain the final high-resolution prediction.}
      \label{fig:sr_pipeline}
\end{figure}

\paragraph{Application to forecasts.}
We apply the super-resolution operator as a post-processing step to coarse-resolution forecast states for all lead times (Figure~\ref{fig:sr_pipeline}a). Given a coarse-resolution forecast trajectory $\left(\hat{\mathbf{x}}_{t+k\delta}^{\mathrm{LR}}\right)_{1 \le k \le T}$, the corresponding high-resolution fields are obtained independently at each lead time as
\begin{equation}
      \hat{\mathbf{x}}_{t+k\delta}^{\mathrm{HR}} =
      \mathcal{C}^{-1}_\phi\!\left(\hat{\mathbf{x}}_{t+k\delta}^{\mathrm{LR}}\right), \; 1 \le k \le T.
\end{equation}

In the following, we consider ArchesWeather and ArchesWeatherGen \citep{couaironArchesWeatherArchesWeatherGenDeterministic2024} as the coarse-resolution forecasting framework, which is augmented by the proposed super-resolution approach.

\section{Experiments}
\subsection{Training and evaluation data}
Training pairs $\left(\mathbf{x}^{\mathrm{LR}},\,\mathbf{x}^{\mathrm{HR}}\right)$ are constructed by regridding ERA5 data at 0.25$^\circ$ resolution to 1.5$^\circ$ using the first-order conservative scheme. This remapping ensures that fluxes remapped to the coarse grid represent the same net energy and mass exchange as on the fine grid \citep{jonesFirstSecondOrderConservative1999}. This setup corresponds to a $6\times$ downscaling task in both latitude and longitude. We use data from 1979--2018 for training, 2019 for validation, and 2020 for testing.

\paragraph{Zero-shot application to forecasts.}
Although the super-resolution model is trained exclusively on reanalysis pairs, we apply it to model forecast outputs in a zero-shot manner. This is possible because ArchesWeatherGen is trained to sample physically plausible coarse-resolution trajectories from the ERA5 data distribution. ArchesWeatherGen produces stochastic ensemble members, such that multiple forecast realizations correspond to the same reanalysis state. As a result, no unique supervised mapping from forecast outputs to ERA5 exists, whereas training exclusively on ERA5-based pairs yields a well-posed inverse regridding task that is independent of forecast model errors.

\subsection{Evaluation methodology}
Our experimental evaluation is organized around two complementary objectives: (i) validating the consistency of the super-resolution operator with respect to the underlying coarse-resolution forecasts, and (ii) assessing the quality of the resulting high-resolution forecasts in a standard benchmarking setting. In both sets of experiments, each high-resolution ensemble member is obtained by super-resolving the corresponding low-resolution ArchesWeatherGen trajectory.

\paragraph{Design validation.}
We first assess the impact of super-resolution on coarse-resolution forecasts produced by ArchesWeatherGen at $1.5^\circ$ resolution.
To this end, we compare re-coarsened super-resolved forecasts against the original ArchesWeatherGen trajectories using pattern correlation, activity ratio, and normalized RMSE.
Re-coarsening is performed using the same conservative regridding operator $\mathcal{C}$ used to construct the coarse-resolution training data.
Pattern correlation is computed as the Pearson correlation coefficient over the latitude-longitude grid, while the activity ratio and normalized RMSE quantify changes in resolved-scale variance and amplitude, respectively. Precise definitions of each metric are provided in Appendix~\ref{app:design_validation_metrics}.

\paragraph{Forecasting benchmark.}
In a second set of experiments, we evaluate the quality of super-resolved forecasts within a standard global weather forecasting framework. Following the WeatherBench~2 protocol \citep{raspWeatherBench2Benchmark2024} and the experimental setup of ArchesWeatherGen, we compare post-processed high-resolution forecasts against established baselines, including IFS ENS \citep{81142}, GenCast \citep{priceProbabilisticWeatherForecasting2025}, and a naive bicubic interpolation of coarse-resolution ArchesWeatherGen forecasts to the target resolution, which serves as the no-added-value baseline. The exact formula for each metric is presented in Appendix~\ref{app:forecasting_metrics}.

\section{Results}
\subsection{Design validation}
\begin{figure}[htb]
      \centering
      \includegraphics[width=\textwidth]{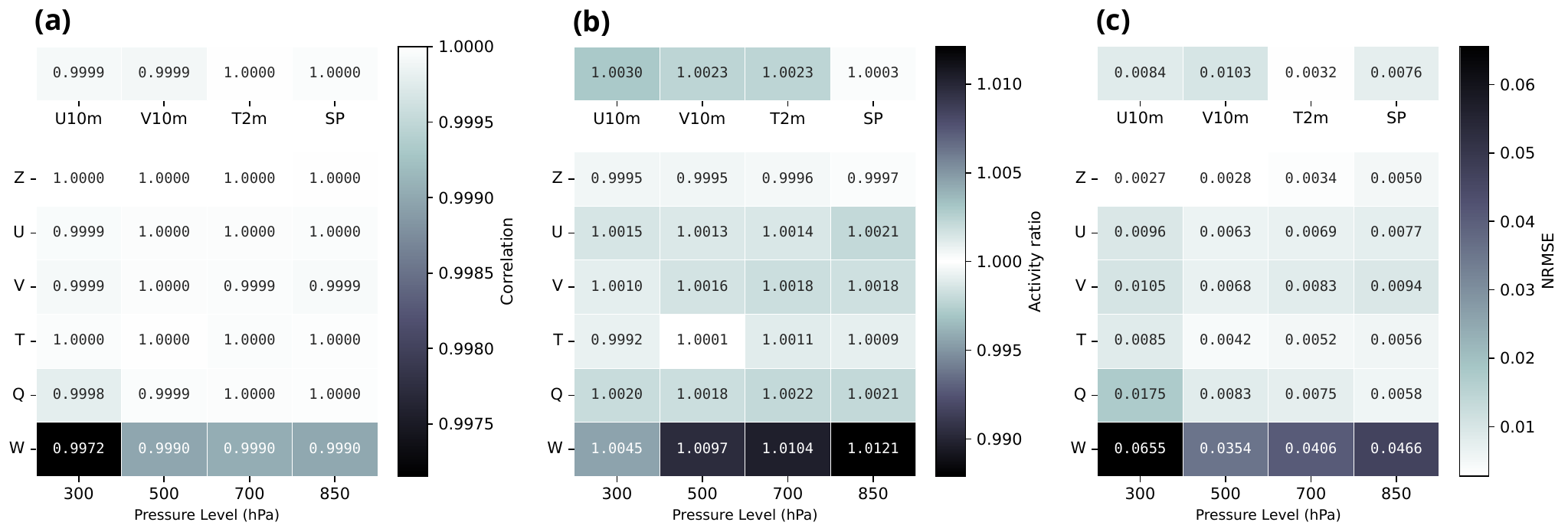}
      \caption{\textbf{Comparison of re-coarsened super-resolved forecasts with the original ArchesWeatherGen coarse-resolution trajectories at 1-day lead time.}
            Shown are (a) spatial correlation, (b) activity ratio, and (c) normalized RMSE, for surface variables and selected pressure levels.
            Color intensity encodes the deviation from the ideal reference: values approaching white indicate near-perfect agreement (unity for correlation and activity ratio, zero for NRMSE), while darker shades correspond to larger departures from the original coarse-resolution trajectories.}
      \label{fig:table_design_validation}
\end{figure}

We assess the consistency of the super-resolution design by comparing re-coarsened super-resolved forecasts to the original coarse-resolution trajectories produced by ArchesWeatherGen at a representative lead time of 1 day (Figure~\ref{fig:table_design_validation}). This table-based view allows a broader coverage of variables and pressure levels while focusing on a representative short-range lead time. Across all considered surface and upper-air variables, spatial correlations remain extremely close to unity, indicating that the large-scale spatial structure is preserved after super-resolution and re-coarsening. Activity ratios stay near one, ruling out both spurious variance injection and excessive smoothing. Normalized RMSE values remain uniformly small, confirming that reinjecting super-resolved states induces only minor, well-controlled deviations from the original coarse trajectories. Taken together, these diagnostics demonstrate that the proposed super-resolution approach leaves the large-scale state essentially unchanged, while introducing only limited and physically consistent subgrid-scale modifications.

\subsection{Quality of super-resolved forecasts}
We now evaluate the quality of super-resolved forecasts using standard ensemble verification metrics as well as physical consistency diagnostics.

\begin{figure}[htb]
      \centering
      \includegraphics[width=0.8\textwidth]{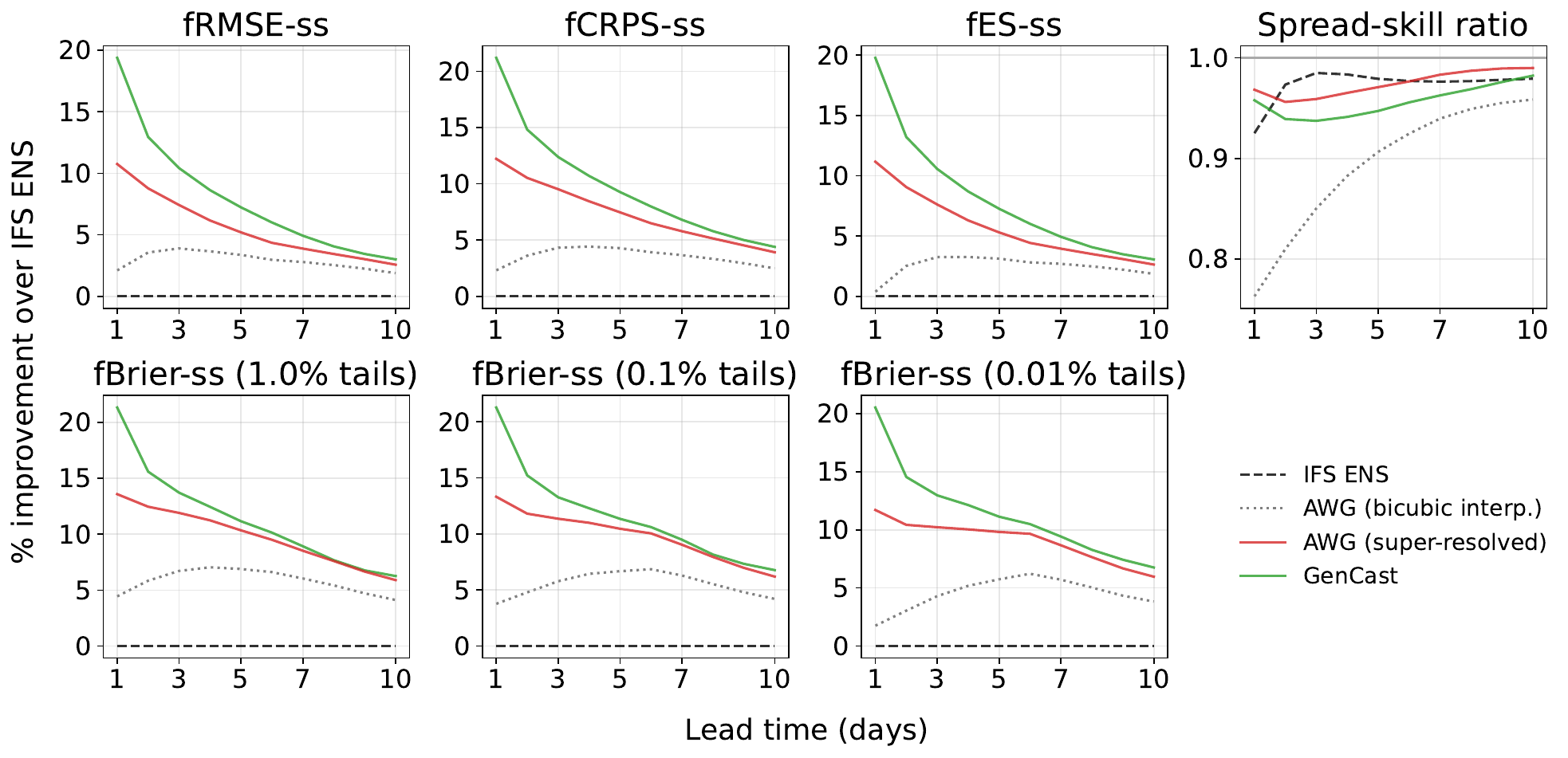}
      \caption{\textbf{Global ensemble forecast skill at 0.25°.}
            Relative improvement over IFS ENS for fair ($f$) ensemble skill-score ($ss$) metrics (Ensemble Mean RMSE, CRPS, Energy Score, Brier Score, spread-skill ratio), averaged over WeatherBench headline variables.}
      \label{fig:global_metrics_eval}
\end{figure}

\paragraph{Summary ensemble metrics.}
Figure~\ref{fig:global_metrics_eval} summarizes the performance of the different models using key ensemble metrics, computed in their fair versions: Ensemble Mean RMSE, CRPS, Energy Score (ES), Brier Score, and the spread-skill ratio. Brier Scores are evaluated at symmetric distribution tails by averaging scores at the corresponding lower and upper quantile thresholds (e.g., 1\% and 99\%). All metrics are expressed as relative improvements with respect to IFS ENS. Scores are averaged over both headline upper-air variables (Z500, Q700, T850, U850, V850) and the four surface variables (T2M, SP, U10M, V10M).

Overall, GenCast exhibits the strongest average performance at short to medium lead times (1--7 days). However, ArchesWeatherGen remains competitive and shows comparable skill at longer lead times (7--10 days), particularly for wind components and specific humidity. This behavior is consistent with the increasing dominance of large-scale dynamics at extended lead times.
In addition, ArchesWeatherGen displays a more favorable spread-skill ratio across all lead times. While the ensemble remains slightly under-dispersive, this under-dispersion is less pronounced than for GenCast.

\begin{figure}[htb]
      \centering
      \includegraphics[width=\textwidth]{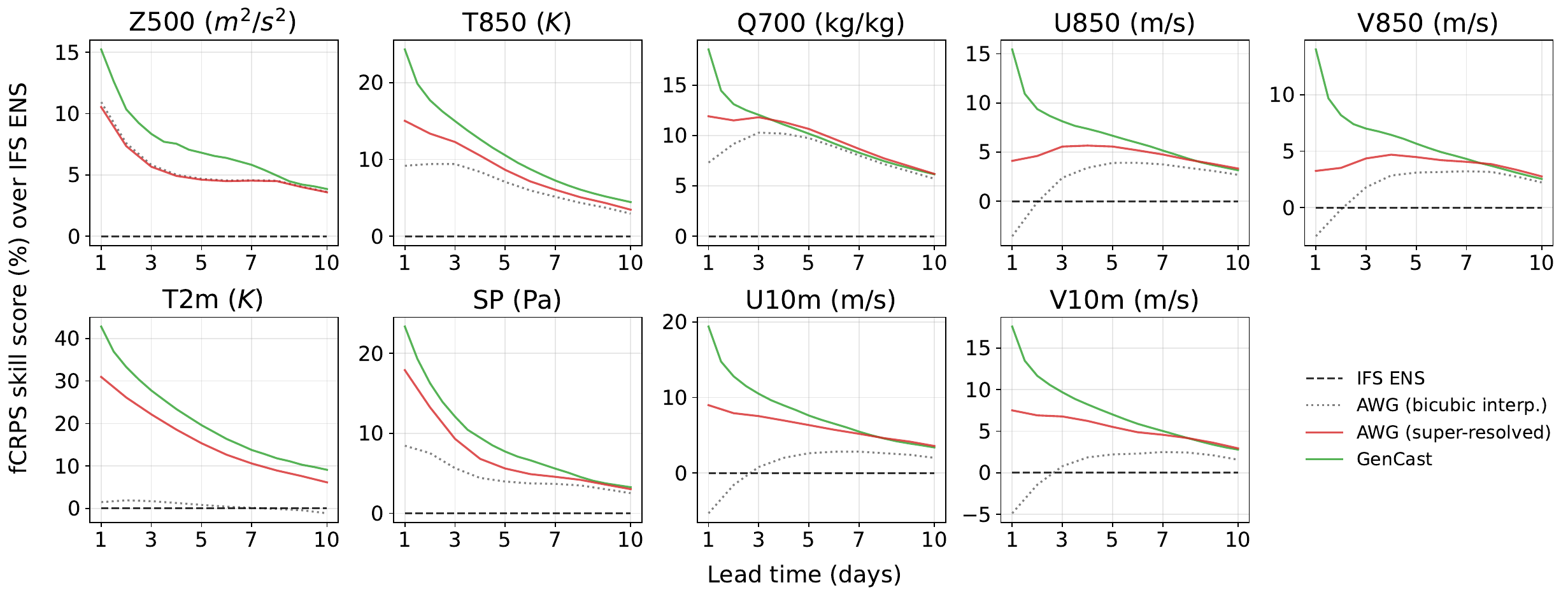}
      \caption{\textbf{Per-variable fair CRPS skill scores relative to IFS ENS at 0.25°.} Results are shown for GenCast, and ArchesWeatherGen (AWG) after bicubic interpolation and learned super-resolution, as a function of lead time.}
      \label{fig:fcrps_skills_score}
\end{figure}

\paragraph{Per-variable CRPS analysis.}
While the previous results focus on metrics averaged across variables, we now examine performance at the level of individual physical variables. Figure~\ref{fig:fcrps_skills_score} reports CRPS skill scores by variable and lead time. We focus on the CRPS, which assesses the quality of the predicted marginal distributions and, when evaluated over long lead times, implicitly penalizes physically inconsistent trajectories in autoregressive models.

ArchesWeatherGen achieves better CRPS skill scores than IFS ENS across all headline variables and lead times, by $7.41\%$ on average. Compared to GenCast, ArchesWeatherGen exhibits slightly worse performance for lead times of 1--6 days, particularly for wind variables, but achieves similar or better CRPS scores at lead times of 7--10 days. A notable exception is specific humidity, for which ArchesWeatherGen converges to GenCast-level performance at earlier lead times (3--10 days).
Relative to bicubic interpolation, the super-resolution approach yields substantially improved probabilistic skill across nearly all variables, with the exception of geopotential. For this variable, bicubic interpolation performs comparatively well due to its intrinsically smooth and large-scale nature. This contrast highlights that the gains from learned super-resolution are most pronounced for variables with richer small-scale variability. We further provide a comparison of fair CRPS skill at 1.5$^\circ$ and 0.25$^\circ$ resolutions in Appendix~\ref{app:crps_scores}.

\begin{figure}[htb]
      \centering
      \includegraphics[width=\textwidth]{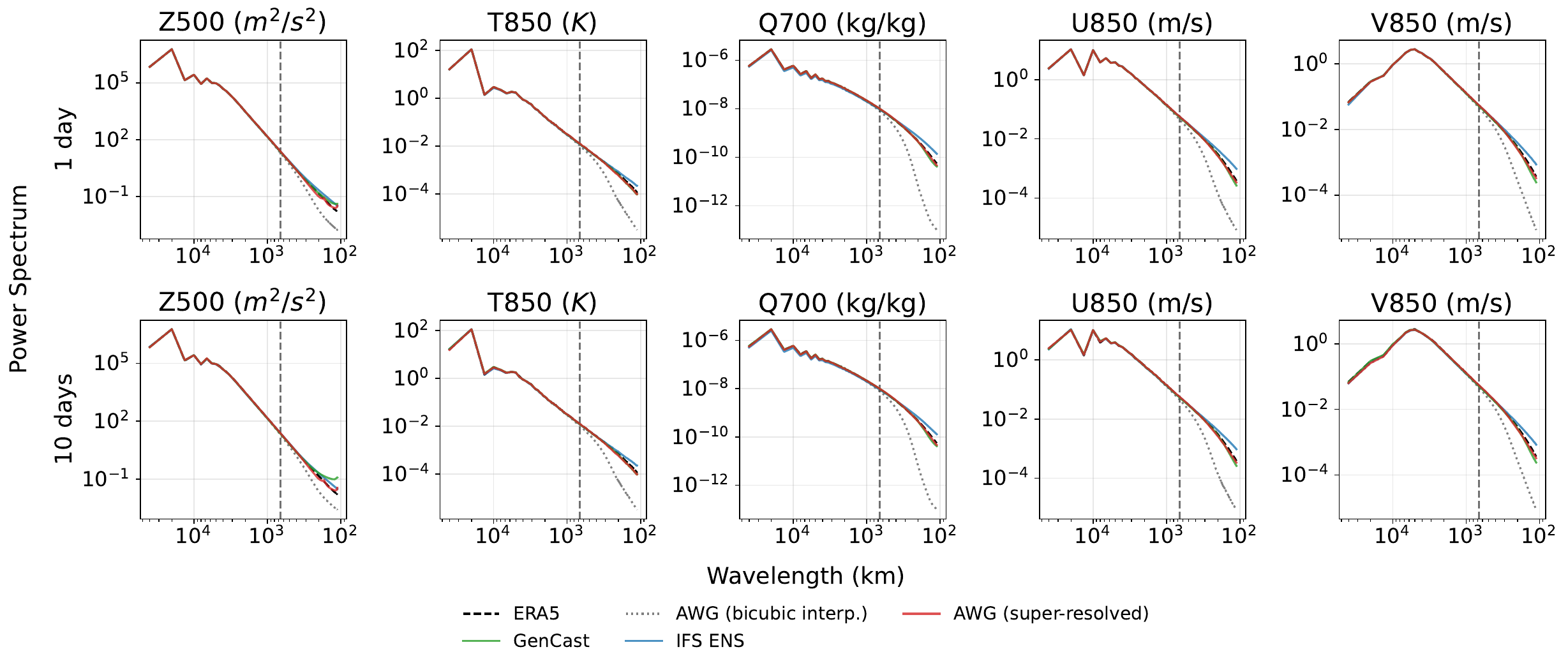}
      \caption{\textbf{Comparison of power spectra across models and lead times.} Power spectra of Z500, T850, Q700, U850, and V850 at 1-day (top) and 10-day (bottom) lead times. For each model and wavelength, the energy at that wavelength is averaged across samples. The dashed vertical line marks the transition between low- and high-resolution scales (1.5° to 0.25°).
      }
      \label{fig:powerspectra}
\end{figure}

\paragraph{Physical consistency and spectral properties.}
Beyond pointwise probabilistic scores, we assess the physical realism of the generated forecasts through power spectral analysis (Figure~\ref{fig:powerspectra}). Power spectra are computed for selected variables at 1-day and 10-day lead times, with energy averaged across ensemble members. The dashed vertical line marks the transition between scales resolved by the low-resolution forecast (1.5°) and those introduced by super-resolution (0.25°).
Both generative models recover substantially more fine-scale energy than bicubic interpolation, demonstrating improved physical consistency at small spatial scales. ArchesWeatherGen exhibits a slightly slower decay of energy at the finest resolved scales compared to GenCast, indicating a better recovery of small-scale variability, although the differences remain moderate.
Overall, the spectral analysis confirms that super-resolved forecasts preserve large-scale structure while consistently recovering fine-scale variability.

\subsection{Case study: Hurricane Teddy}
\begin{figure}[htb]
      \centering
      \includegraphics[width=\textwidth]{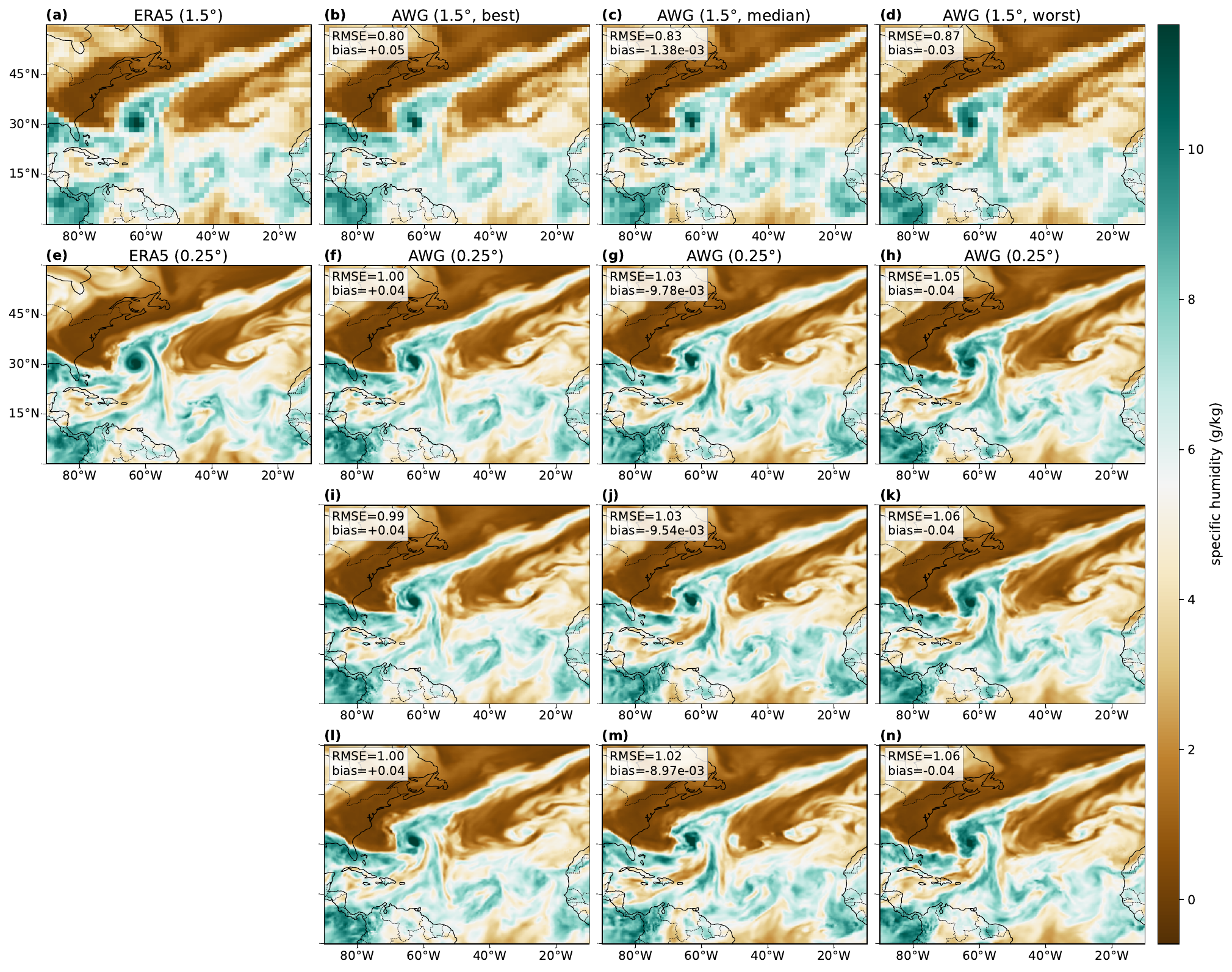}
      \caption{\textbf{Super-resolution of specific humidity at 700 hPa over the North Atlantic (Hurricane Teddy, 21 September 2020 12:00 UTC, +24\,h lead time).} ERA5 regridded to 1.5° \textbf{(a)} and at native 0.25° \textbf{(e)} serve as low- and high-resolution ground truth respectively. AWG ensemble forecasts at 1.5° are shown for the best, median, and worst members by RMSE \textbf{(b–d)}. Each column of the high-resolution block \textbf{(f,i,l)}, \textbf{(g,j,m)}, \textbf{(h,k,n)} shows three independent stochastic SR draws at 0.25° for the corresponding LR member. RMSE and bias (g\,kg$^{-1}$) against the appropriate ERA5 reference are reported in the top-left corner of each panel.}
      \label{fig:q700_teddy}
\end{figure}

To complement the quantitative evaluation, we present a case study of Hurricane Teddy, a major Atlantic storm that reached Category 4 intensity in September 2020. It produced one of the most organized deep convective structures of that season. At its peak, Teddy exhibited a well-defined eye, vigorous spiral rainbands, and a broad moisture envelope that spanned much of the subtropical North Atlantic. Figure~\ref{fig:q700_teddy} showcases ArchesWeatherGen super-resolved humidity fields at a 1-day lead time, initialized on September 20, 2020.

At 1.5° resolution, the low-resolution ensemble members (b--d) accurately capture the large-scale moisture distribution and the hurricane's position, aligning closely with the ERA5 reference (a). Ensemble spread is visible in the representation of the cyclone core: it is less well-defined in the worst-performing member (d) than in the median (c) and best (b) members, while the synoptic-scale flow and storm placement remain consistent across all members.
Super-resolution is then applied to each low-resolution member (f--n). At 0.25°, the SR model reconstructs fine-scale structures including cyclonic moisture spirals, indications of a dry intrusion on the western flank, and sharper outer gradients. For each member, multiple stochastic draws produce distinct realizations that differ in the positioning of small-scale moisture filaments, while preserving the large-scale organization inherited from the conditioning low-resolution state. Importantly, this behavior holds consistently across lead times: Appendix~\ref{app:case_study_teddy_7d} shows that analogous fine-scale structures are recovered at 7-day lead time, demonstrating that the SR model produces physically plausible reconstructions throughout the forecast horizon.
It should be noted that such stochastic variability does not necessarily improve forecast skill. Since the SR model is conditioned on the low-resolution input, large-scale errors are not corrected and propagate to the super-resolved outputs. Probabilistic scores may therefore penalize the ensemble if it is misaligned with the reference, despite exhibiting realistic fine-scale variability. Nevertheless, by representing physically plausible subgrid structures, super-resolution enables the exploration of internal variability and provides a more complete depiction of uncertainty at fine scales, which is particularly relevant for intense systems such as tropical cyclones.
To further quantitatively evaluate the model's capacity beyond open-ocean dynamics, we extend the analysis to a topographically complex region during the passage of Storm Alex over the Alps (Appendix~\ref{app:case_study_alex}).

\section{Discussion}
We investigated learned super-resolution as a modular extension to global weather forecasting models trained at coarse spatial resolution. By decoupling forecasting and downscaling, our approach enables the reconstruction of physically plausible fine-scale structure while preserving the large-scale dynamics learned at low resolution, without retraining forecasting models at operational resolution. Despite its modest computational cost, this strategy achieves competitive forecast performance relative to substantially more expensive end-to-end high-resolution forecasting models.

\paragraph{Generality and scope.}
Although demonstrated using the ArchesWeatherGen framework, the proposed approach is not tied to a specific forecasting architecture. In principle, it can be applied to any forecasting model provided that (i) its coarse-resolution outputs represent physically realistic large-scale flow and (ii) its temporal sampling is consistent with the dynamics it resolves.
In this work, we focused on daily (24\,h) forecast trajectories, a setting in which the method reliably reconstructs fine-scale spatial structure while preserving the underlying coarse-resolution forecast trajectory at the resolved time scale. Extending such daily trajectories to sub-daily (e.g., hourly) outputs would require a dedicated temporal super-resolution component, which we leave for future work.
Moreover, by building on probabilistic forecasts rather than deterministic forecasts, the approach naturally decouples large- and fine-scale uncertainties, allowing each to be modeled and sampled independently. Extending the method to deterministic, smoothed inputs would instead require learning a direct or unpaired mapping from forecast outputs to high-resolution reference data \citep{wanDebiasCoarselySample2023,keislerSerpentFlowGenerativeUnpaired2026}.

\paragraph{Broader implications.}
Beyond medium-range weather forecasting, our framework naturally extends to subseasonal and climate applications, where large-scale dynamics are generally well represented but fine-scale variability remains under-resolved. In this context, super-resolution models could be trained on high-resolution climate simulations or reanalysis-derived datasets and applied in a zero-shot manner to coarse global climate models, as demonstrated in related work \citep{hessFastScaleadaptiveUncertaintyaware2025}. A particularly promising avenue is the use of global or limited-area kilometer-scale models, such as ICON \citep{hoheneggerICONSapphireSimulatingComponents2023}, as sources of high-resolution reference data.

\paragraph{Efficiency and limitations.}
A key advantage of the proposed approach is computational efficiency. Training the full pipeline—deterministic forecasting, generative forecasting, and super-resolution—amounts to on the order of 30 A100-days in total. Super-resolution training represents a relatively small fraction of this cost, requiring approximately 7 A100-days, compared to about 23 A100-days for training ArchesWeatherGen.
At inference time, generating a full 10-day high-resolution forecast takes about 8 minutes on a single NVIDIA V100 32GB GPU, which can become significant when applied at scale, such as for long test sets or large ensembles. This computational cost motivates future work on speeding up inference through distillation or consistency-based generative modeling \citep{stockSwiftAutoregressiveConsistency2025}.

\paragraph{Conclusion.}
Overall, this work demonstrates that learned super-resolution provides a physically grounded and computationally efficient pathway towards higher-resolution weather and climate prediction. By separating large-scale prediction from fine-scale reconstruction, the proposed framework offers flexibility, interpretability, and scalability across forecasting horizons and application domains. Our results demonstrate that this approach achieves competitive forecast skill at operational resolution with substantially reduced computational cost, and that recovered small-scale variability exhibits realistic spectral characteristics.

{\small
\paragraph{\small Acknowledgments.}
The authors thank Guillaume Couairon, David Landry, and Pierre Garcia for fruitful discussions.
This project was granted HPC resources by GENCI at IDRIS under the allocation AD011016443 on the Jean-Zay supercomputer.

\paragraph{\small Data Availability Statement.}
The ERA5 reanalysis data used in this study are publicly available from the Copernicus Climate Data Store (\url  {https://doi.org/10.24381/cds.adbb2d47}). The code used in this study is available at \url{https://doi.org/10.5281/zenodo.19355356}.
}

\printbibliography

\newpage
\appendix
\markboth{APPENDIX}{APPENDIX}
\section{Training details}
\subsection{ERA5}
We use a subset of the ERA5 reanalysis from the WeatherBench~2 dataset, selected to exactly match the input and output configuration of ArchesWeatherGen. The dataset includes the same set of atmospheric variables, comprising upper-air fields on 13 pressure levels (50, 100, 150, 200, 250, 300, 400, 500, 600, 700, 850, 925, and 1000~hPa) together with surface variables (Table~\ref{tab:variables}), all provided on a regular latitude-longitude grid.
Data are sampled at 6-hour intervals (00, 06, 12, and 18~UTC), consistent with the temporal resolution used during model training.
ERA5 fields at 0.25$^\circ$ resolution are regridded to 1.5$^\circ$ using the first-order conservative scheme, resulting in the coarse-resolution ERA5 data used to train ArchesWeatherGen.
Input coarse-resolution states and residuals are normalized separately, using per-variable and per-level statistics computed from ERA5 over the 1979--2018 training period:
\begin{equation}
      \tilde{\mathbf{x}}^{\mathrm{LR}} =
      \frac{\mathbf{x}^{\mathrm{LR}} - \mu_{\mathbf{x}^{\mathrm{LR}}}}{\sigma_{\mathbf{x}^{\mathrm{LR}}}},
      \qquad
      \tilde{\mathbf{r}} =
      \frac{\mathbf{r} - \mu_{\mathbf{r}}}{\sigma_{\mathbf{r}}}.
\end{equation}

\begin{table}[htbp]
      \centering
      \begin{tabular}{lll}\toprule
            {Type}      & {Variable name}              & {Short name} \\\midrule
            Atmospheric & Geopotential                 & $Z$          \\
            Atmospheric & Specific humidity            & $Q$          \\
            Atmospheric & Temperature                  & $T$          \\
            Atmospheric & U component of wind          & $U$          \\
            Atmospheric & V component of wind          & $V$          \\
            Atmospheric & Vertical velocity            & $W$          \\
            Surface     & 2 metre temperature          & $T2m$        \\
            Surface     & Mean sea-level pressure      & $SP$         \\
            Surface     & 10 metre U component of wind & $U10m$       \\
            Surface     & 10 metre V component of wind & $V10m$       \\
            Static      & Geopotential height          & $Z$          \\
            Static      & Land-sea mask                & LSM          \\
            Static      & Soil-type mask               & STM          \\
            Time        & Month of year                & ---          \\
            Time        & Hour of day                  & ---          \\
            \bottomrule
      \end{tabular}\vspace{1em}
      \caption{\textbf{ECMWF data variables used in our datasets.} \emph{Atmospheric} represents time-varying atmospheric properties and \emph{Surface} represents time-varying surface-level properties. \emph{Static} and \emph{time} variables are used as additional conditioning for the model.}\label{tab:variables}
\end{table}

\subsection{Model architecture and training}\label{app:model_architecture}
\begin{figure}[ht]
      \centering
      \includegraphics[width=0.7\textwidth]{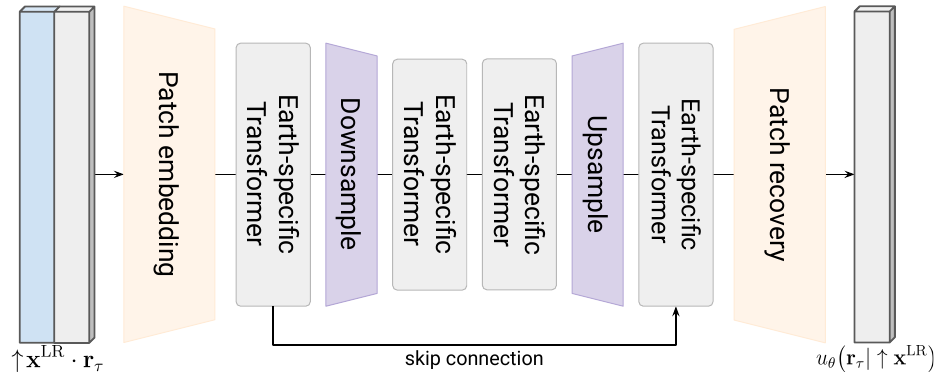}
      \caption{\textbf{Architecture of the 3D Swin U-Net.} The interpolated coarse-resolution state $\uparrow\!\mathbf{x}^{\mathrm{LR}}$ and the noisy residual $\mathbf{r}_\tau$ are concatenated and passed through the network, predicting the velocity field $u_\theta\left(\mathbf{r}_\tau\!\mid\,\uparrow\!\mathbf{x}^{\mathrm{LR}}\right)$.}
      \label{fig:backbone}
\end{figure}

Here we provide additional details on the conditioning mechanism, training objective, and architectural hyperparameters.
\paragraph{Conditioning.} The coarse-resolution state $\uparrow\!\mathbf{x}^{\mathrm{LR}}$ is bicubically interpolated to the target resolution and concatenated with the noisy residual $\mathbf{r}_\tau$ along the variable dimension before being passed to the network, following the SR3 framework \citep{sahariaImageSuperResolutionIterative2023}. The resulting input is processed by a 3D Swin U-Net following ArchesWeather \citep{couaironArchesWeatherArchesWeatherGenDeterministic2024}, whose architecture is illustrated in Figure~\ref{fig:backbone}.

\paragraph{Training objective.} The model is trained using flow matching \citep{lipmanFlowMatchingGenerative2023}, which learns to transport samples from a standard Gaussian prior toward the data distribution by regressing a velocity field along linear interpolation paths between noise and data. Concretely, given a clean residual $\mathbf{r}_0$ and noise $\boldsymbol{\epsilon} \sim \mathcal{N}(0, I)$, a noisy intermediate $\mathbf{r}_\tau = \alpha_\tau \mathbf{r}_0 + \sigma_\tau \boldsymbol{\epsilon}$ is constructed at timestep $\tau$, and the network $u_\theta\left(\mathbf{r}_\tau\!\mid\,\uparrow\!\mathbf{x}^{\mathrm{LR}}\right)$ is trained to predict the target velocity $\mathbf{r}_0 - \boldsymbol{\epsilon}$. The noise schedule $(\alpha_\tau, \sigma_\tau)$ follows the sigmoid of a normal distribution as in Stable Diffusion~3 \citep{esserScalingRectifiedFlow2024}.

\paragraph{Loss weighting.} Following GraphCast \citep{lamLearningSkillfulMediumrange2023}, the training loss is weighted to account for both the spherical geometry of the Earth and the vertical structure of the atmosphere. Latitude-longitude area weighting is applied to correct for grid-cell area distortion, and air-density-proportional coefficients are used to weight pressure levels. However, we apply a uniform weighting across physical variables, assigning a coefficient of 1 to all variables.
We also considered adding an explicit constraint penalizing deviations in large-scale structure after re-coarsening, but did not pursue this, as the desired behavior emerged with the baseline loss formulation.

\paragraph{Hyperparameters.} All architectural hyperparameters are identical to ArchesWeatherGen, with the sole exception of the patch embedding size, which is increased from $[2,2,2]$ to $[2,4,4]$. The model is trained using the AdamW optimizer with a learning rate of $3 \times 10^{-4}$, $\beta_1 = 0.9$, $\beta_2 = 0.98$, and a weight decay of $0.05$. We use a batch size of 4 and train for 75{,}000 optimization steps on four NVIDIA A100 80GB GPUs, corresponding to approximately 40 hours of training.

\section{Evaluation metrics}\label{app:metrics}

In the following, we provide the definitions of all statistical estimators used as evaluation metrics. We recall that we consider weather states consisting of six upper-air variables (temperature, geopotential, specific humidity, wind components $U$, $V$ and $W$) sampled on a latitude-longitude grid at 13 pressure levels, and four surface variables (2m temperature, mean sea-level pressure, and 10m wind components $U$ and $V$).

\subsection{Design validation metrics}
\label{app:design_validation_metrics}

First, we present a set of diagnostic metrics designed to assess whether super-resolution modifies the large-scale atmospheric state when re-coarsened to the original forecast resolution. These metrics are used exclusively for design validation and aim to verify that the super-resolution operator preserves large-scale structure and variance, without introducing spurious artifacts.

Let $\mathbf{x}^{\mathrm{LR}, v}_{t,\delta}$ denote the original coarse-resolution forecast at lead time $\delta$, for variable $v$, and $\tilde{\mathbf{x}}^{\mathrm{SR},v}_{t,\delta}$ the corresponding super-resolved field after re-coarsening to the same latitude-longitude grid.

\paragraph{Pattern correlation.}
Pattern correlation measures the similarity of large-scale spatial structures between two fields and is defined as the Pearson correlation coefficient evaluated over the latitude-longitude grid:
\begin{equation}
      \mathrm{Corr}(v,\delta)
      =
      \frac{
      \left\langle
      \tilde{\mathbf{x}}^{\mathrm{SR},v}_{t,\delta}
      -
      \overline{\tilde{\mathbf{x}}^{\mathrm{SR},v}_{t,\delta}},
      \;
      \mathbf{x}^{\mathrm{LR},v}_{t,\delta}
      -
      \overline{\mathbf{x}^{\mathrm{LR},v}_{t,\delta}}
      \right\rangle
      }{
      \sqrt{
            \left\langle
            \left(
            \tilde{\mathbf{x}}^{\mathrm{SR},v}_{t,\delta}
            -
            \overline{\tilde{\mathbf{x}}^{\mathrm{SR},v}_{t,\delta}}
            \right)^{2}
            \right\rangle
            \left\langle
            \left(
            \mathbf{x}^{\mathrm{LR},v}_{t,\delta}
            -
            \overline{\mathbf{x}^{\mathrm{LR},v}_{t,\delta}}
            \right)^{2}
            \right\rangle
      }
      }.
\end{equation}
Here, $\langle \cdot \rangle$ denotes the spatial average over the latitude-longitude grid, and the overline indicates the spatial mean. A value close to $1$ indicates that large-scale spatial patterns are preserved after super-resolution and re-coarsening.

\paragraph{Activity and activity ratio.}
To quantify changes in spatial variability, we compute the activity, defined as the spatial variance of anomaly fields relative to a climatological mean. Let $\mathbf{c}^{v}_{t+\delta}$ denote the climatology for variable $v$ at the corresponding time of year. The activity is defined as
\begin{equation}
      \mathrm{Activity}(v,\delta)
      =
      \frac{1}{T}
      \sum_{t}
      \left\langle
      \left(
      \mathbf{x}^{v}_{t,\delta}
      -
      \mathbf{c}^{v}_{t+\delta}
      -
      \overline{\mathbf{x}^{v}_{t,\delta}
            -
            \mathbf{c}^{v}_{t+\delta}}
      \right)^{2}
      \right\rangle.
\end{equation}
We report the activity ratio between re-coarsened super-resolved fields and the original coarse-resolution forecasts:
\begin{equation}
      \mathrm{ActivityRatio}(v,\delta)
      =
      \frac{
            \mathrm{Activity}\!\left(\tilde{\mathbf{x}}^{\mathrm{SR},v}_{t,\delta}\right)
      }{
            \mathrm{Activity}\!\left(\mathbf{x}^{\mathrm{LR},v}_{t,\delta}\right)
      }.
\end{equation}
An activity ratio close to $1$ indicates that super-resolution does not inject spurious variance or induce excessive smoothing at resolved scales.

\paragraph{Normalized RMSE.}
Finally, we compute a normalized Root Mean Square Error between the re-coarsened super-resolved fields and the original coarse-resolution forecasts:
\begin{equation}
      \mathrm{NRMSE}(v,\delta)
      =
      \sqrt{
      \frac{
      \frac{1}{T}
      \sum_{t}
      \left\|
      \tilde{\mathbf{x}}^{\mathrm{SR},v}_{t,\delta}
      -
      \mathbf{x}^{\mathrm{LR},v}_{t,\delta}
      \right\|_{2}^{2}
      }{
      \sigma^{2}_{\mathrm{ERA5}}(v)
      }
      }.
\end{equation}
Here, $\sigma_{\mathrm{ERA5}}(v)$ denotes the climatological standard deviation of variable $v$ estimated from the ERA5 training period. Normalization allows RMSE values to be compared across variables with different physical units. A value close to $0$ indicates that re-coarsened super-resolved fields remain nearly identical to the original coarse-resolution forecasts.

\subsection{Forecasting metrics}
\label{app:forecasting_metrics}
We present the metrics used to evaluate the quality of probabilistic forecasts, following the WeatherBench~2 evaluation protocol \citep{raspWeatherBench2Benchmark2024}. We report the fair versions of these estimators \citep{ferroFairScoresEnsemble2014}, which are explicitly defined below. These fair estimators correspond to the theoretical values obtained in the limit of an infinite ensemble size, and therefore allow metrics to be compared consistently across different ensemble sizes.
All metrics are computed with latitude weighting to account for the area distortion induced by the latitude-longitude grid, such that the contribution of each grid cell is proportional to its surface area on the sphere.

For a given initialization time $t$, and a forecast model that produces an ensemble of $M$ forecasts, we denote $\hat{\mathbf{x}}^{v,i}_{t,\delta}$ the forecast of the $i$-th ensemble member at lead time $\delta$, for variable $v$. The ground truth target is $\mathbf{x}^{v}_{t+\delta}$.

\paragraph{Ensemble Mean RMSE.}
The ensemble mean is defined as the average of the ensemble members:
\begin{equation}
      \bar{\mathbf{x}}^{v}_{t,\delta}
      =
      \frac{1}{M}
      \sum_{i=1}^{M}
      \hat{\mathbf{x}}^{v,i}_{t,\delta}.
\end{equation}
The fair Ensemble Mean RMSE is then defined as
\begin{equation}
      \mathrm{FairEnsMeanRMSE}(v,\delta)
      =
      \sqrt{
            \frac{1}{T}
            \sum_{t}\left(
            \left\|
            \bar{\mathbf{x}}^{v}_{t,\delta}
            -
            \mathbf{x}^{v}_{t+\delta}
            \right\|_{2}^{2}
            -
            \frac{1}{M(M-1)}
            \sum_{i=1}^{M}
            \left\|
            \hat{\mathbf{x}}^{v,i}_{t,\delta}
            -
            \bar{\mathbf{x}}^{v}_{t,\delta}
            \right\|_{2}^{2}
            \right)
      }.
\end{equation}

\paragraph{CRPS and Energy Score.}
The Continuous Ranked Probability Score (CRPS) assesses the quality of marginal predictive distributions. Its fair estimator is defined as
\begin{equation}
      \mathrm{CRPS}(v,\delta)
      =
      \frac{1}{T}
      \sum_{t}
      \left(
      \frac{1}{M}
      \sum_{i=1}^{M}
      \left\|
      \hat{\mathbf{x}}^{v,i}_{t,\delta}
      -
      \mathbf{x}^{v}_{t+\delta}
      \right\|_{1}
      -
      \frac{1}{2M(M-1)}
      \sum_{i,j}
      \left\|
      \hat{\mathbf{x}}^{v,i}_{t,\delta}
      -
      \hat{\mathbf{x}}^{v,j}_{t,\delta}
      \right\|_{1}
      \right).
\end{equation}
The Energy Score (ES) is defined analogously by replacing the $\ell_{1}$ norm with the $\ell_{2}$ norm. As a multivariate generalization of the CRPS, it evaluates the joint spatial and cross-variable consistency of ensemble forecasts, whereas the CRPS only assesses marginal (univariate) distributions.

\paragraph{Brier Score.}
Let $c^{v}_{q}$ denote the $q$-quantile of the climatology for variable $v$. The empirical exceedance probability is
\begin{equation}
      p^{v}_{t,\delta,q}
      =
      \frac{1}{M}
      \sum_{i=1}^{M}
      \mathds{1}
      \!\left(
      \hat{\mathbf{x}}^{v,i}_{t,\delta}
      >
      c^{v}_{q}
      \right).
\end{equation}
The fair Brier Score is then defined as
\begin{equation}
      \mathrm{FairBrierScore}(v,\delta,q)
      =
      \frac{1}{T}
      \sum_{t}
      \left(
      \left\|
      p^{v}_{t,\delta,q}
      -
      \mathds{1}
      \!\left(
      \mathbf{x}^{v}_{t+\delta}
      >
      c^{v}_{q}
      \right)
      \right\|_{2}^{2}
      -
      \frac{1}{M-1}
      p^{v}_{t,\delta,q}
      \left(
      1 - p^{v}_{t,\delta,q}
      \right)
      \right).
\end{equation}

\paragraph{Spread-Skill Ratio.}
The spread of an ensemble is defined as the square root of its unbiased variance estimator
\begin{equation}
      \mathrm{Spread}(v,\delta)
      =
      \sqrt{
      \frac{1}{T}
      \sum_{t}
      \frac{1}{M-1}
      \sum_{i=1}^{M}
      \left\|
      \hat{\mathbf{x}}^{v,i}_{t,\delta}
      -
      \bar{\mathbf{x}}^{v}_{t,\delta}
      \right\|_{2}^{2}
      }.
\end{equation}
The spread-skill ratio is then
\begin{equation}
      \mathrm{SSR}(v,\delta)
      =
      \sqrt{\frac{M+1}{M}}
      \,
      \frac{
            \mathrm{Spread}(v,\delta)
      }{
            \mathrm{EnsMeanRMSE}(v,\delta)
      }.
\end{equation}
A value close to $1$ indicates a well-calibrated ensemble; values below (above) $1$ indicate under- (over-) dispersion.

\paragraph{Skill scores.}
To compare models against a reference forecast, we report skill scores defined as relative improvements. For instance, the CRPS skill score is defined as
\begin{equation}
      \mathrm{CRPS\text{-}SS}_{\mathrm{model}}(v,\delta)
      =
      1
      -
      \frac{
      \mathrm{CRPS}_{\mathrm{model}}(v,\delta)
      }{
      \mathrm{CRPS}_{\mathrm{ref}}(v,\delta)
      }.
\end{equation}
Skill scores are dimensionless and can therefore be averaged across variables:
\begin{equation}
      \mathrm{CRPS\text{-}SS}_{\mathrm{model}}(\delta)
      =
      \frac{1}{\|\mathcal{V}\|}
      \sum_{v \in \mathcal{V}}
      \mathrm{CRPS\text{-}SS}_{\mathrm{model}}(v,\delta).
\end{equation}

\section{Additional experiments}
\subsection{Design validation}
\begin{figure}[ht]
      \centering
      \includegraphics[width=\textwidth]{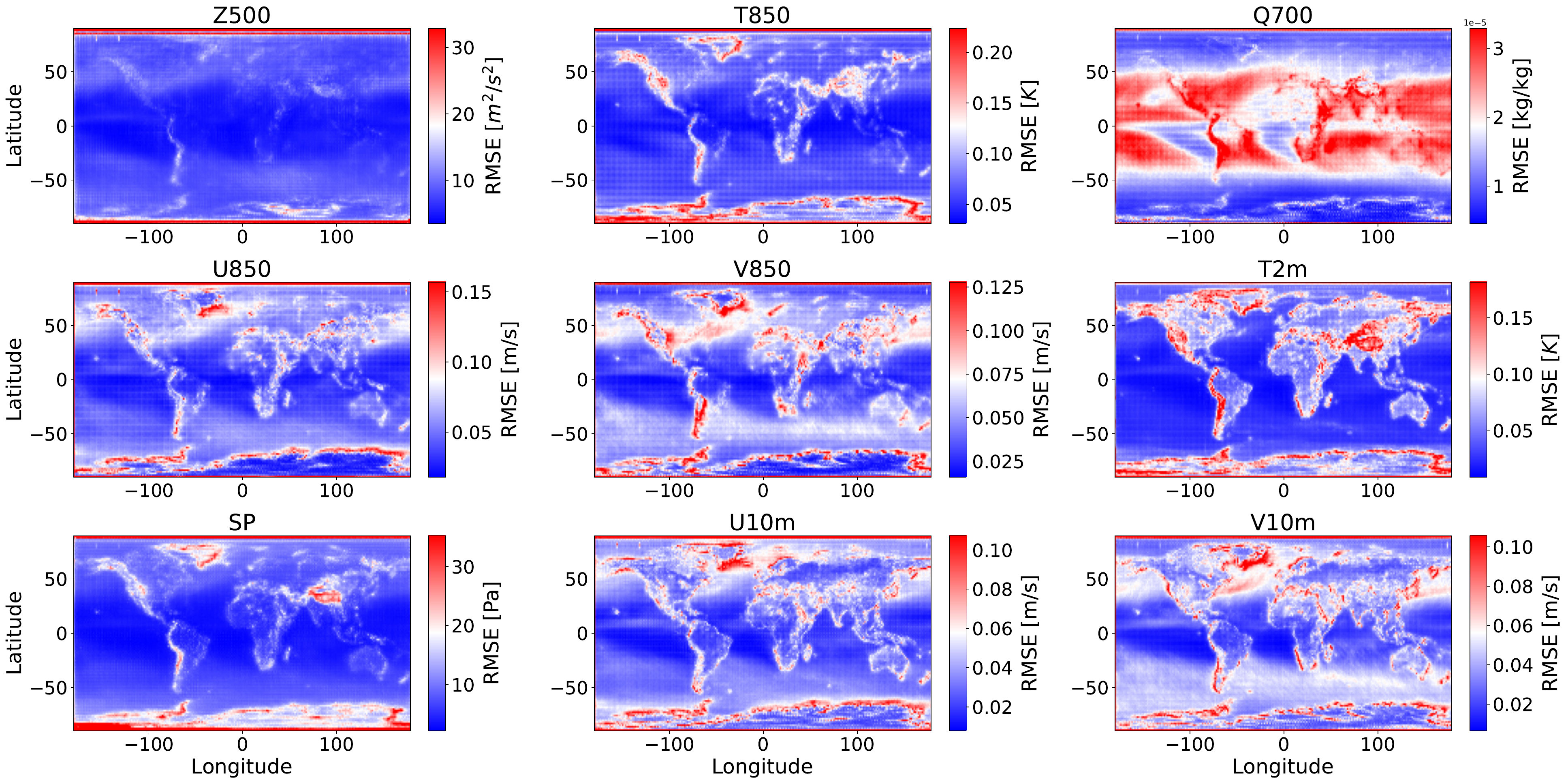}
      \caption{\textbf{Super-resolution error at 1-day lead time averaged over the 2020 test period.}
            Maps of RMSE computed between re-coarsened super-resolved forecasts and the original ArchesWeatherGen trajectories for selected upper-air and surface variables.}
      \label{fig:diff_rmse_maps}
\end{figure}

As an extension of the design validation experiments presented in the main text, we further examine the spatial structure of differences introduced by super-resolution (Figure~\ref{fig:diff_rmse_maps}).
Differences, although small, are spatially localized and primarily concentrated in regions characterized by strong gradients and complex surface forcing, such as mountainous areas. These regions are particularly sensitive to subgrid-scale processes, and the observed patterns indicate that post-processed super-resolution mainly affects locations where unresolved variability is dynamically relevant, while leaving the large-scale structure elsewhere largely unchanged.

\subsection{Preservation of ensemble skill under super-resolution}\label{app:crps_scores}

\begin{figure}[ht]
      \centering
      \includegraphics[width=\textwidth]{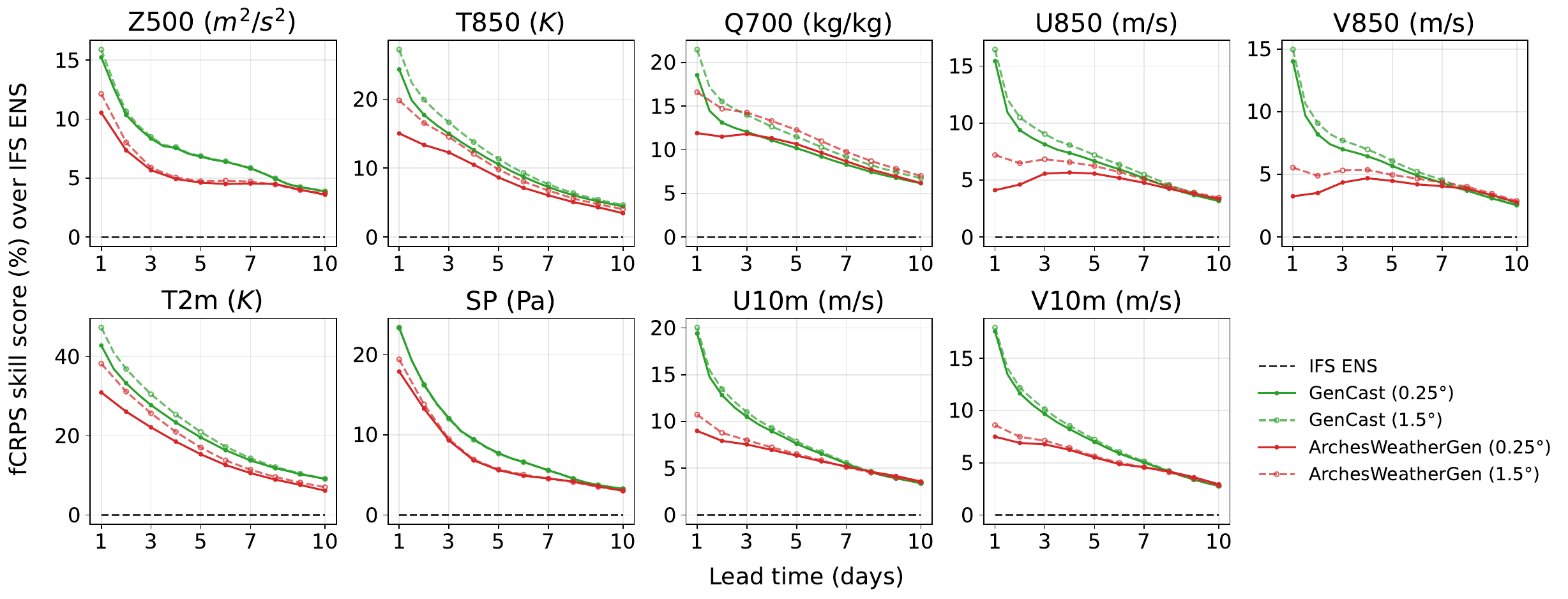}
      \caption{
            \textbf{Per-variable fair CRPS skill scores relative to IFS ENS at the corresponding resolution}. Results are shown for ArchesWeatherGen and GenCast as a function of lead time at both 1.5$^\circ$ and 0.25$^\circ$ resolutions. ArchesWeatherGen forecasts are produced natively at 1.5$^\circ$ and super-resolved to 0.25$^\circ$, while GenCast forecasts are produced natively at 0.25$^\circ$ and conservatively regridded to 1.5$^\circ$.
      }
      \label{fig:fcrps_skills_score_1.5deg_with_gencast}
\end{figure}

Figure~\ref{fig:fcrps_skills_score_1.5deg_with_gencast} compares fair CRPS skill scores of ensemble forecasts from ArchesWeatherGen and GenCast at 1.5$^\circ$ and 0.25$^\circ$ resolutions, relative to IFS ENS (evaluated at the corresponding resolutions). ArchesWeatherGen forecasts are produced natively at 1.5$^\circ$ resolution and subsequently super-resolved to 0.25$^\circ$, while GenCast forecasts are produced natively at 0.25$^\circ$ resolution and conservatively regridded to 1.5$^\circ$.
Across all headline variables, forecasts at 0.25$^\circ$ and 1.5$^\circ$ exhibit similar qualitative evolution of fair CRPS skill with lead time for both models. For most variables, slightly higher skill is observed at 1.5$^\circ$ at short lead times (days~1--3), while the skill scores at the two resolutions converge at longer lead times (4--10 days), consistent with the higher intrinsic predictability of large-scale atmospheric structures and the reduced impact of small-scale uncertainty. For specific humidity, forecasts at 1.5$^\circ$ consistently exhibit higher skill than at 0.25$^\circ$ for all lead times, reflecting the low predictability of moisture at small spatial scales \citep{kalnayAtmosphericModelingData2003}. Overall, these results indicate that increasing the output resolution to 0.25$^\circ$---either through super-resolution (ArchesWeatherGen) or native high-resolution generation (GenCast)---does not lead to systematic improvements in probabilistic forecast skill relative to the corresponding lower-resolution forecasts.

\subsection{Quality of super-resolved forecasts}

\begin{figure}[ht]
      \centering
      \includegraphics[width=\textwidth]{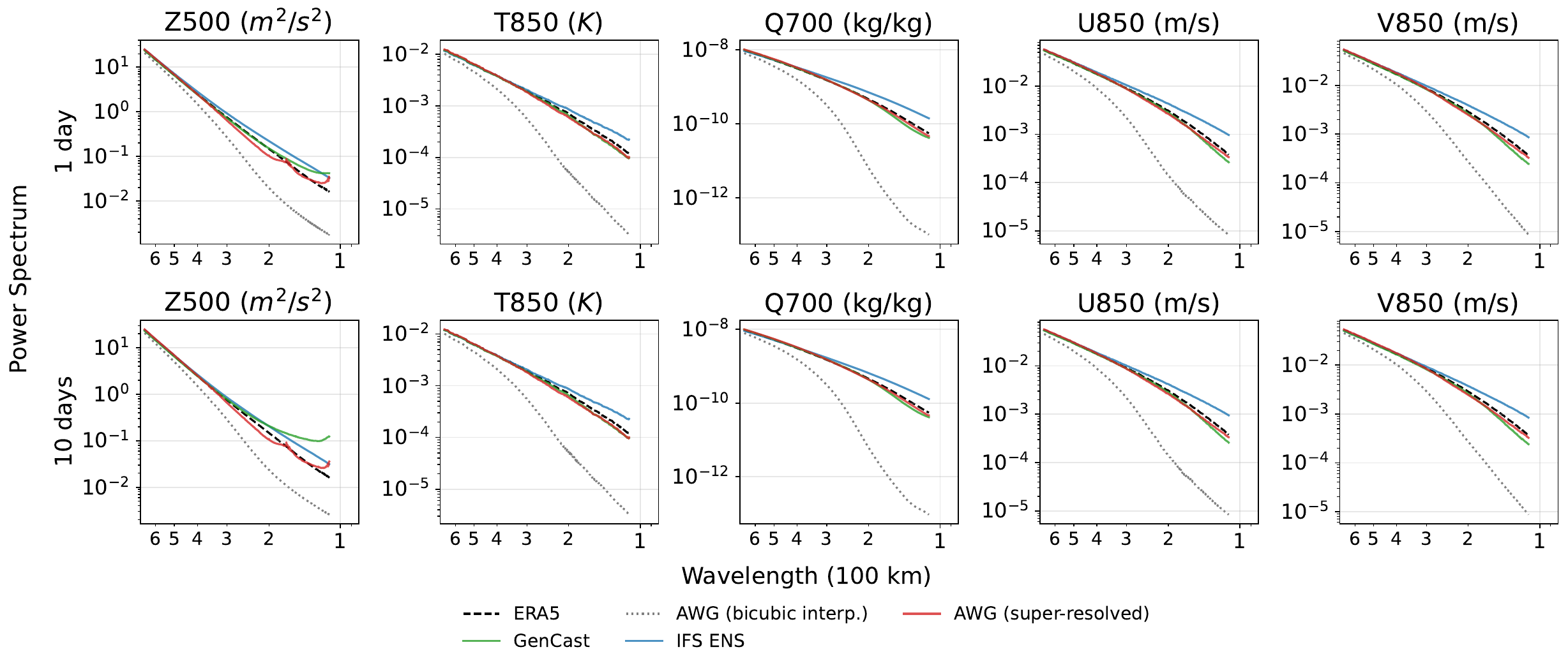}
      \caption{\textbf{Zoomed power spectra at unresolved scales.}
            At scales finer than 1.5$^\circ$, super-resolved ArchesWeatherGen (AWG) exhibits closer agreement with ERA5 compared to GenCast, indicating a more consistent representation of small-scale energy.}
      \label{fig:powerspectra_clean_zoom}
\end{figure}

\begin{figure}[ht]
      \centering
      \includegraphics[width=\textwidth]{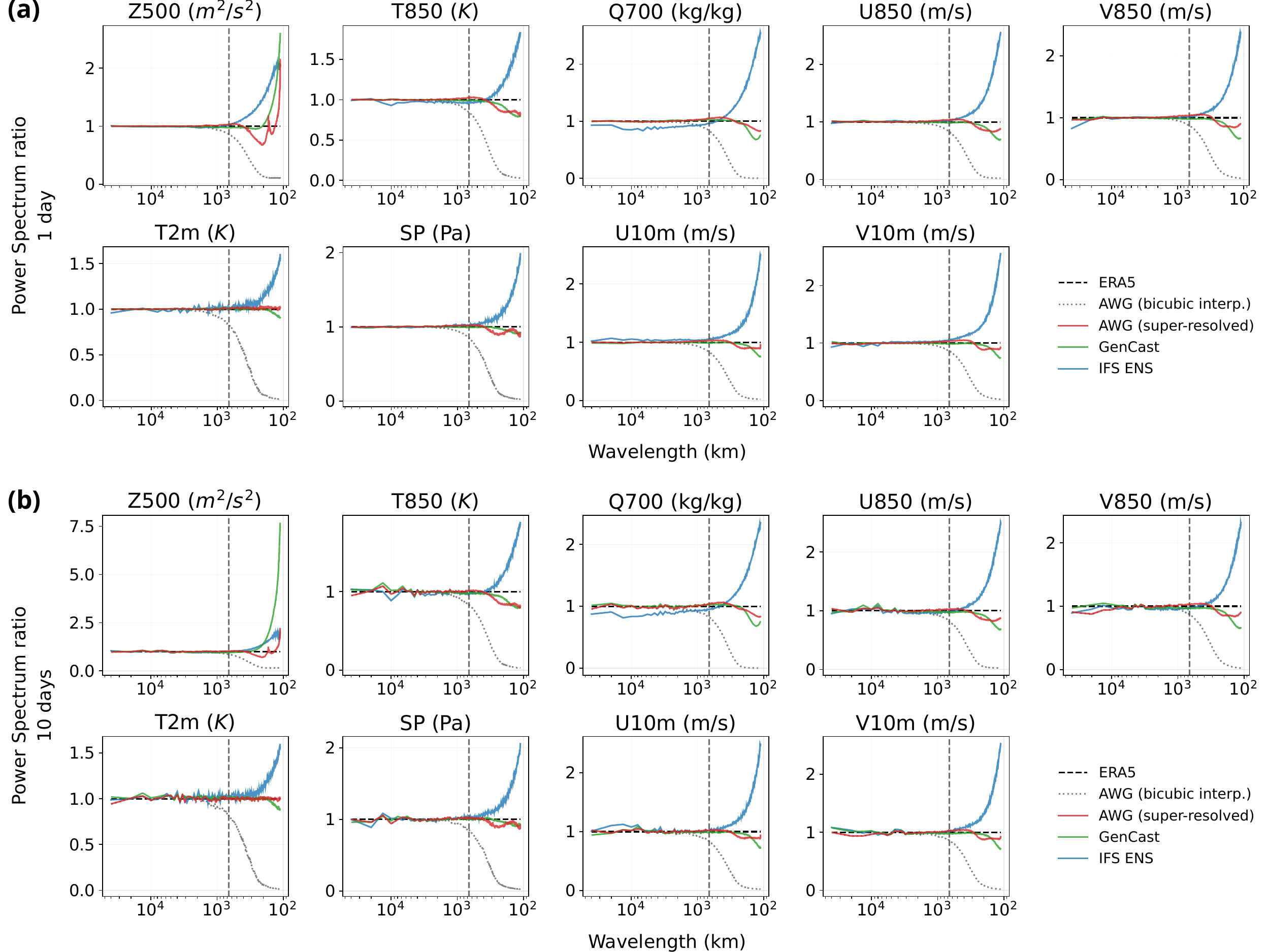}
      \caption{\textbf{Power spectrum ratios relative to ERA5.}
            Ratio of power spectra at 0.25$^\circ$ resolution relative to ERA5 for selected variables, shown as a function of wavelength, at (a) 1-day and (b) 10-day lead times. The vertical dashed line indicates the coarse-resolution cutoff at 1.5$^\circ$.
            A ratio close to unity indicates good agreement with ERA5 at the corresponding scale.}
      \label{fig:ps_ratio_0.25deg_full}
\end{figure}

\paragraph{Additional spectral diagnostics.}
To further assess the reconstruction of small-scale structures, we provide complementary spectral analyses focusing on wavelengths finer than the coarse grid spacing (i.e., scales smaller than 1.5$^\circ$). Figure~\ref{fig:powerspectra_clean_zoom} shows a zoomed view of the rightmost part of the power spectra, isolating the unresolved scales targeted by super-resolution.
We additionally report ratios of power spectra relative to ERA5 at 0.25$^\circ$ resolution for lead times of 1 and 10 days (Figure~\ref{fig:ps_ratio_0.25deg_full}). This facilitates a quantitative comparison across models. Across variables and lead times, ArchesWeatherGen exhibits more consistent small-scale energy compared to GenCast, with particularly strong performance for 2m temperature, while bicubic interpolation systematically underestimates fine-scale variance, as expected.

\subsection{Case study: Hurricane Teddy at 7-day lead time}\label{app:case_study_teddy_7d}

\begin{figure}[htb]
      \centering
      \includegraphics[width=\textwidth]{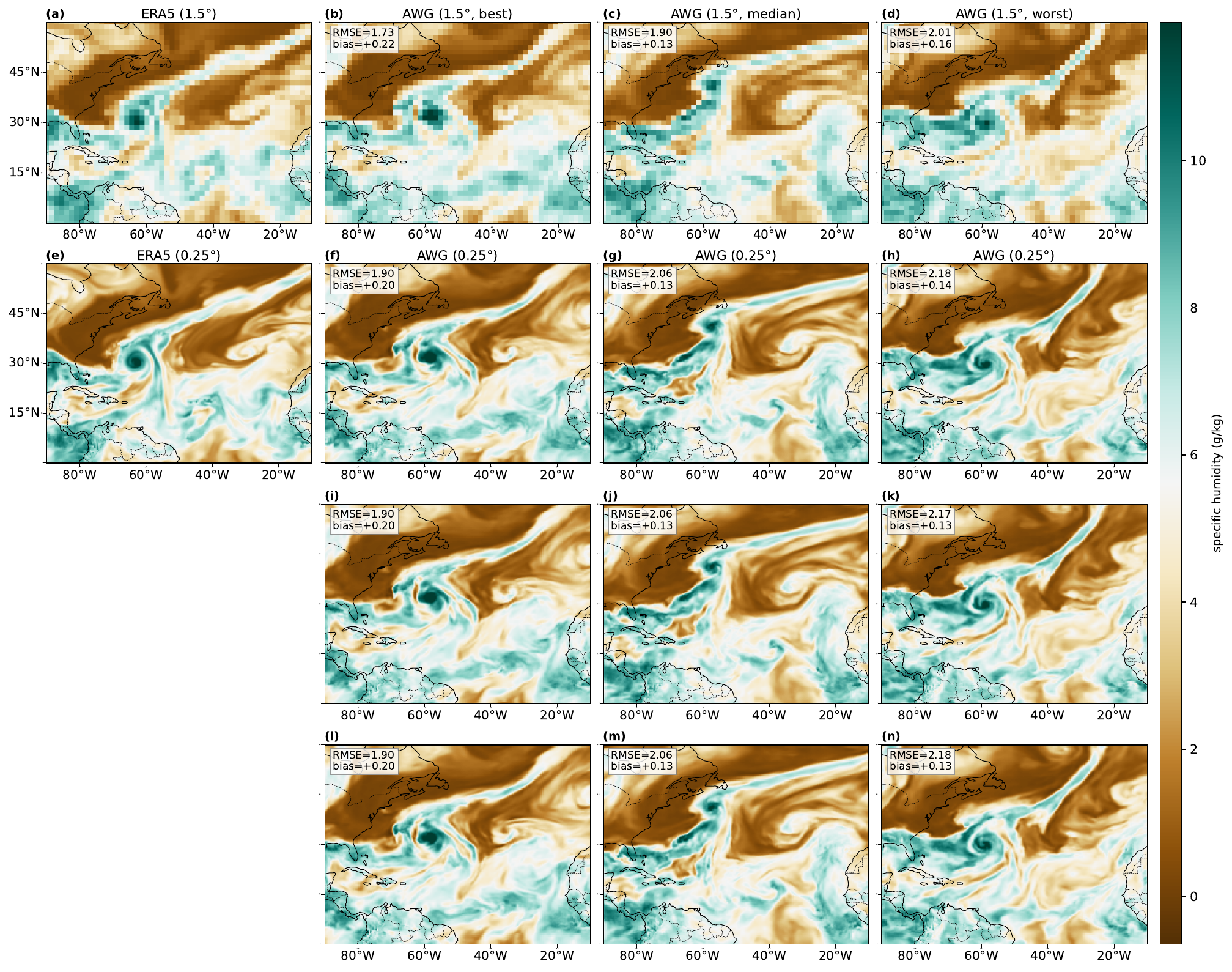}
      \caption{\textbf{Super-resolution of specific humidity at 700 hPa over the North Atlantic (Hurricane Teddy, 21 September 2020 12:00 UTC, +7 day lead time).} ERA5 regridded to 1.5° \textbf{(a)} and at native 0.25° \textbf{(e)} serve as low- and high-resolution ground truth respectively. AWG ensemble forecasts at 1.5° are shown for the best, median, and worst members by RMSE \textbf{(b–d)}. Each column of the high-resolution block \textbf{(f,i,l)}, \textbf{(g,j,m)}, \textbf{(h,k,n)} shows three independent stochastic SR draws at 0.25° for the corresponding LR member. RMSE and bias (g\,kg$^{-1}$) against the appropriate ERA5 reference are reported in the top-left corner of each panel.}
      \label{fig:q700_teddy_7d}
\end{figure}

Figure~\ref{fig:q700_teddy_7d} presents super-resolved specific humidity fields for Hurricane Teddy at a 7-day lead time, initialized on September 14, 2020. At this extended range, the low-resolution ensemble members exhibit increased spread in both storm position and intensity, reflecting the growth of forecast uncertainty over time. Despite this, the super-resolution model consistently reconstructs physically coherent fine-scale structures for each member, including cyclonic moisture organization and sharp outer gradients, conditioned on the respective low-resolution state. This confirms that the SR model does not degrade at longer lead times and produces plausible subgrid-scale variability regardless of forecast uncertainty inherited from the coarse-resolution ensemble.

\subsection{Case study: Storm Alex}\label{app:case_study_alex}

We present a case study centered on extratropical cyclone Alex, which caused exceptional flooding across the Mediterranean region, particularly in the south of France and the Franco-Italian border area. We focus on a reduced orographic domain encompassing the western Mediterranean and the Alps, where the interaction between synoptic-scale dynamics and complex terrain is particularly pronounced.

\begin{figure}[htb]
      \centering
      \includegraphics[width=\textwidth]{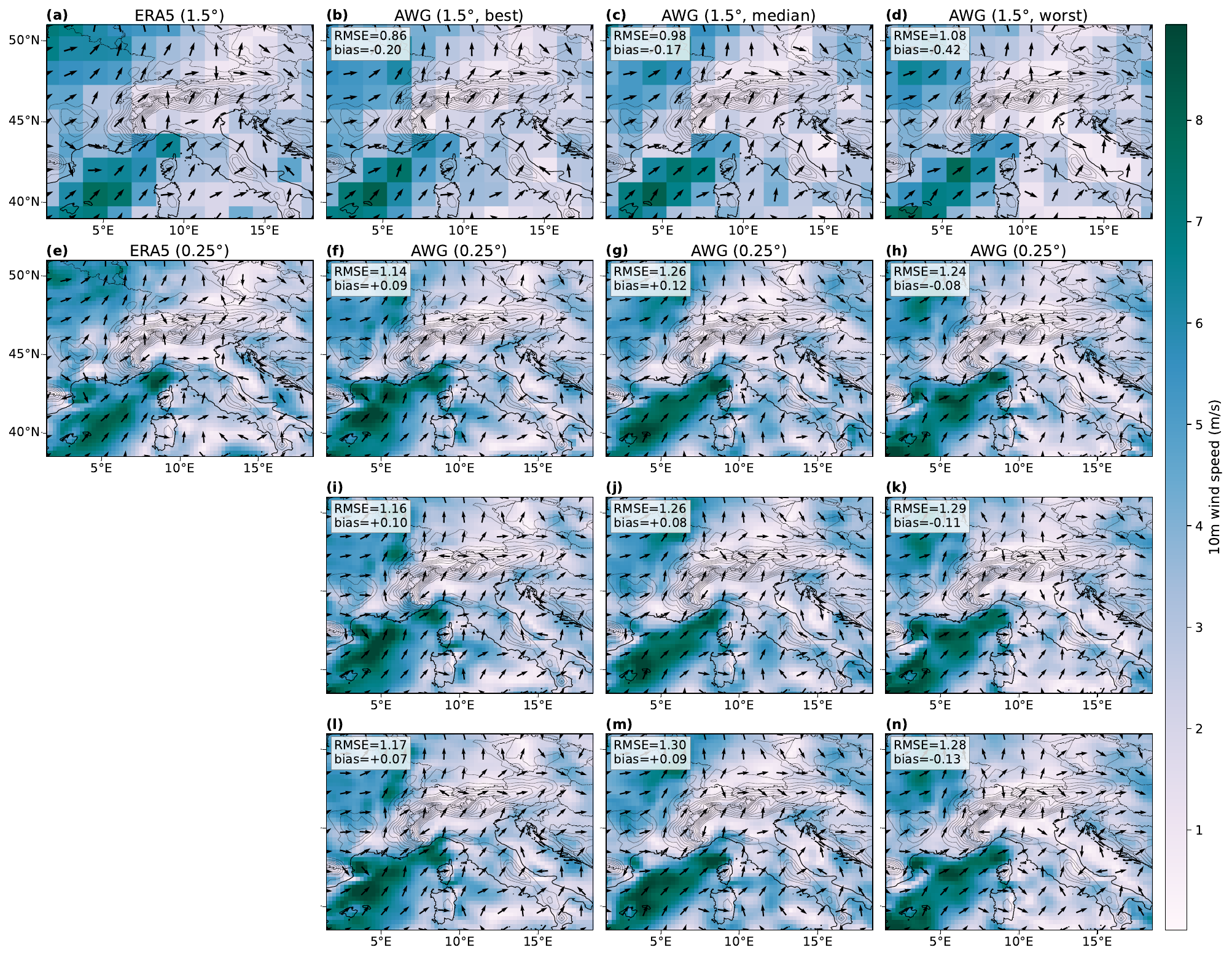}
      \caption{\textbf{Super-resolution of 10m wind speed over the western Mediterranean and Alps (Extratropical Cyclone Alex, 2020-09-30 12:00 UTC, +24\,h lead time).} ERA5 regridded to 1.5° \textbf{(a)} and at 0.25° \textbf{(e)} serve as low- and high-resolution ground truth respectively. AWG ensemble forecasts at 1.5° are shown for the best, median, and worst members by RMSE \textbf{(b--d)}. Each column of the high-resolution block \textbf{(f,i,l)}, \textbf{(g,j,m)}, \textbf{(h,k,n)} shows three independent stochastic SR draws at 0.25° for the corresponding LR member. Wind speed is derived as $\|\mathbf{u}\| = \sqrt{u^2 + v^2}$ from the zonal and meridional wind components; arrows indicate the local wind direction. RMSE and bias (m\,s$^{-1}$) against the appropriate ERA5 reference are reported in the top-left corner of each panel.}
      \label{fig:10m_wind_speed_alps_alex}
\end{figure}

\begin{figure}[htb]
      \centering
      \includegraphics[width=\textwidth]{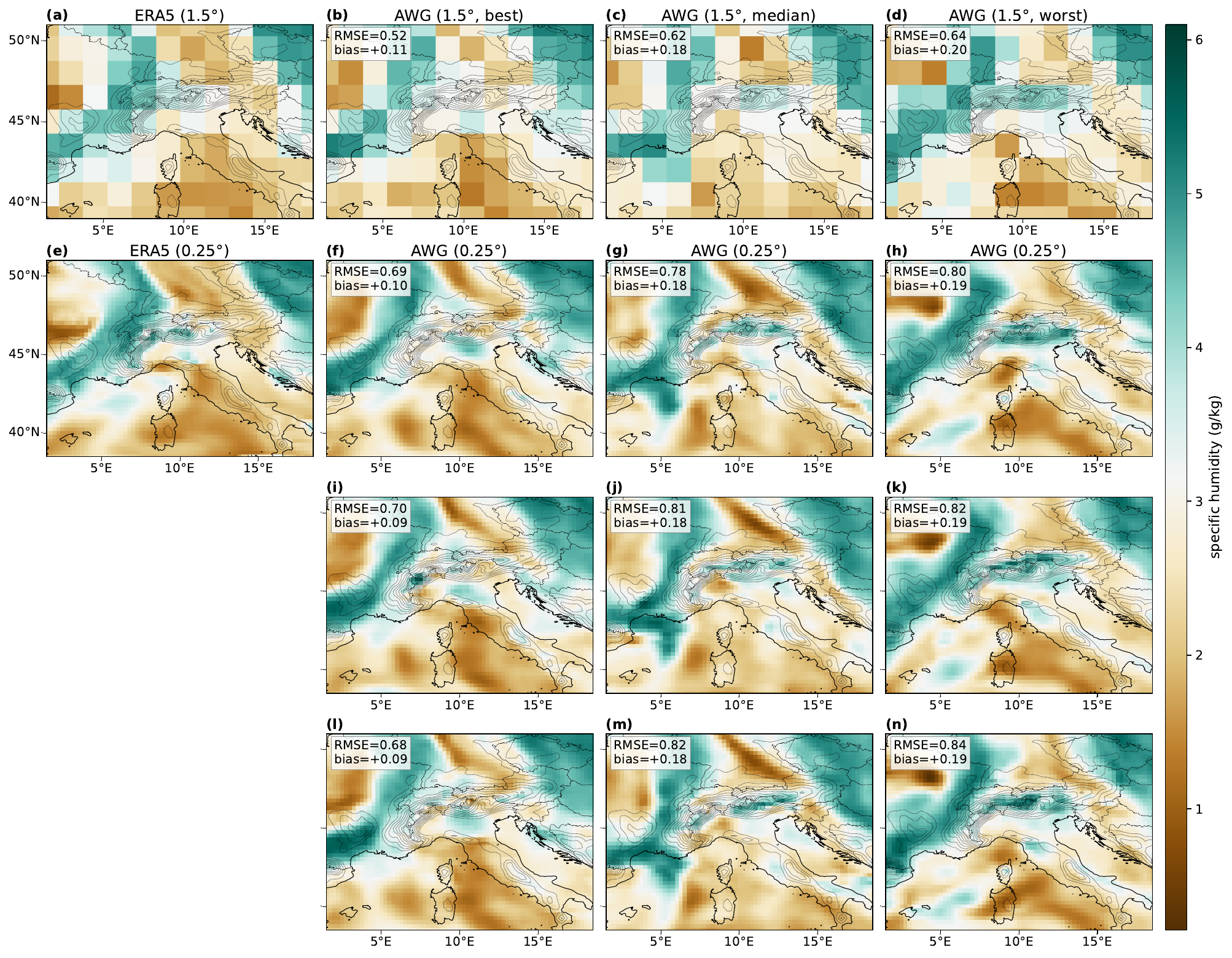}
      \caption{\textbf{Super-resolution of specific humidity at 700 hPa over the western Mediterranean and Alps (Extratropical Cyclone Alex, 2020-09-30 12:00 UTC, +24\,h lead time).} ERA5 regridded to 1.5° \textbf{(a)} and at native 0.25° \textbf{(e)} serve as low- and high-resolution ground truth respectively. AWG ensemble forecasts at 1.5° are shown for the best, median, and worst members by RMSE \textbf{(b--d)}. Each column of the high-resolution block \textbf{(f,i,l)}, \textbf{(g,j,m)}, \textbf{(h,k,n)} shows three independent stochastic SR draws at 0.25° for the corresponding LR member. RMSE and bias (g\,kg$^{-1}$) against the appropriate ERA5 reference are reported in the top-left corner of each panel.}
      \label{fig:q700_alps_alex}
\end{figure}

\begin{figure}[htb]
      \centering
      \includegraphics[width=\textwidth]{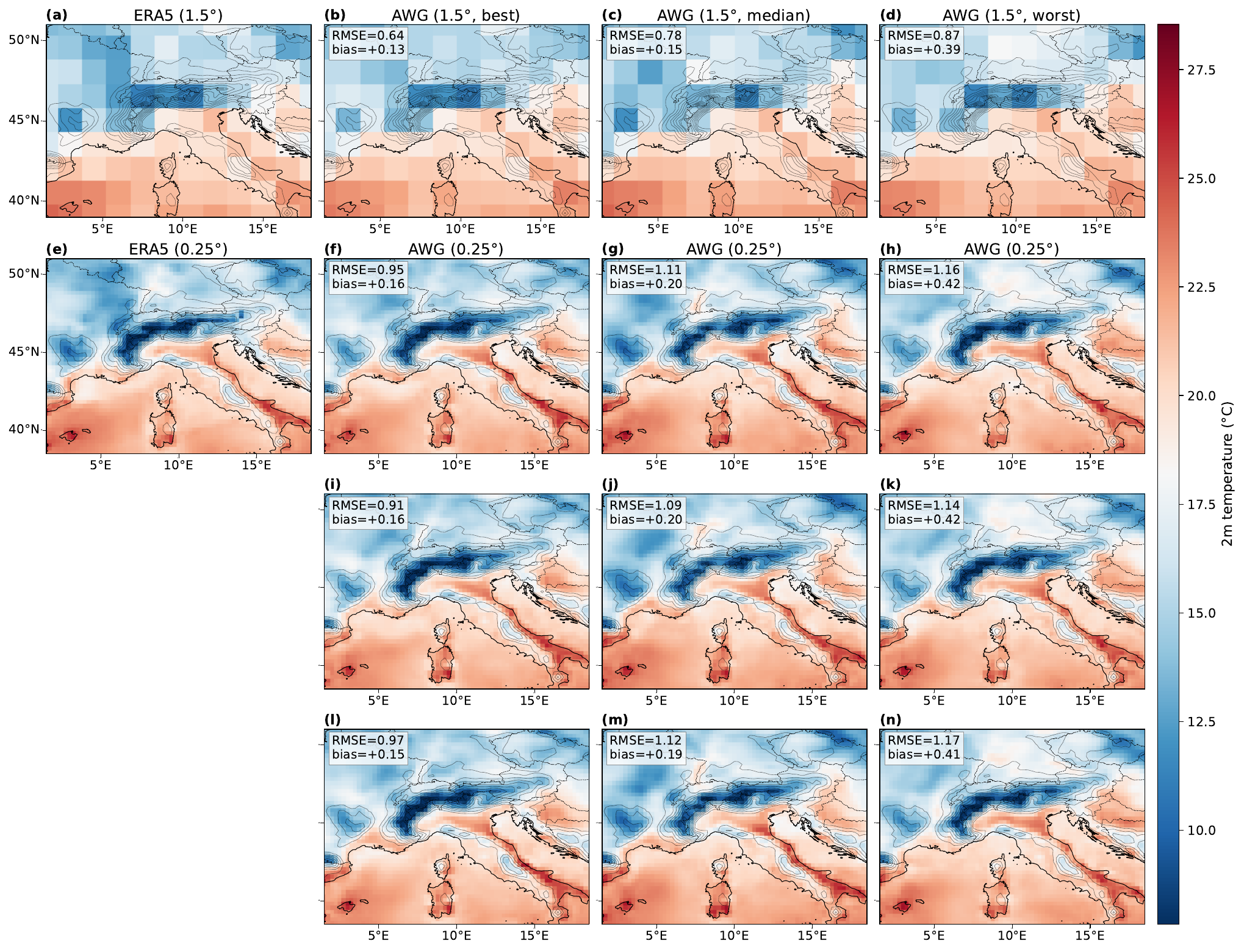}
      \caption{\textbf{Super-resolution of 2m temperature over the western Mediterranean and Alps (Extratropical Cyclone Alex, 2020-09-30 12:00 UTC, +24\,h lead time).} ERA5 regridded to 1.5° \textbf{(a)} and at native 0.25° \textbf{(e)} serve as low- and high-resolution ground truth respectively. AWG ensemble forecasts at 1.5° are shown for the best, median, and worst members by RMSE \textbf{(b--d)}. Each column of the high-resolution block \textbf{(f,i,l)}, \textbf{(g,j,m)}, \textbf{(h,k,n)} shows three independent stochastic SR draws at 0.25° for the corresponding LR member. RMSE and bias (°C) against the appropriate ERA5 reference are reported in the top-left corner of each panel.}
      \label{fig:t2m_alps_alex}
\end{figure}

Forecasts are initialized at 2020-09-30 12:00 UTC, corresponding to the early development stage of the cyclone as it tracked northeastward toward southern Brittany, ahead of the exceptional precipitation episode that would affect the Alpes-Maritimes two days later. We examine three meteorological fields at a 1-day lead time: 10m wind speed (derived as $\|\mathbf{u}\| = \sqrt{u^2 + v^2}$ from the zonal and meridional wind components, as our model does not predict wind speed directly), specific humidity at 700 hPa, and 2m temperature (Figures~\ref{fig:10m_wind_speed_alps_alex}, \ref{fig:q700_alps_alex}, and~\ref{fig:t2m_alps_alex}, respectively). The joint analysis of wind speed and lower-tropospheric specific humidity provides a useful proxy for assessing atmospheric conditions that are conducive to flooding \citep{meyerAtmosphericConditionsFavouring2022}. The 2m temperature field, for which our model achieves particularly strong reconstruction skill, serves as an additional qualitative benchmark.

Across all three fields, the model preserves the large-scale synoptic structure of the input forecast while recovering fine-scale features absent from it. For wind speed, this includes orographic channeling and acceleration along Alpine valleys and the Ligurian coast. For specific humidity, the model sharpens the moisture front by recovering the contrast between dry conditions over the Tyrrhenian Sea and the moist air mass over the Alpine region. For 2m temperature, the sharp thermal gradient along the Alpine arc is consistently reconstructed across members.
Inter-member differences are most pronounced for wind speed and specific humidity, where members exhibit clear variations in the positioning and intensity of topographic features, while 2m temperature shows more consistency across members. This reflects the higher intrinsic small-scale variability of wind and moisture fields, and highlights the added value of a generative approach over a deterministic one.

\section{Pipeline-integrated super-resolution}
In addition to the post-processing application studied in the main text, we also investigated an alternative way of applying super-resolution within the forecasting workflow, which we now describe.

\begin{figure}[ht]
      \centering
      \includegraphics[width=0.5\textwidth]{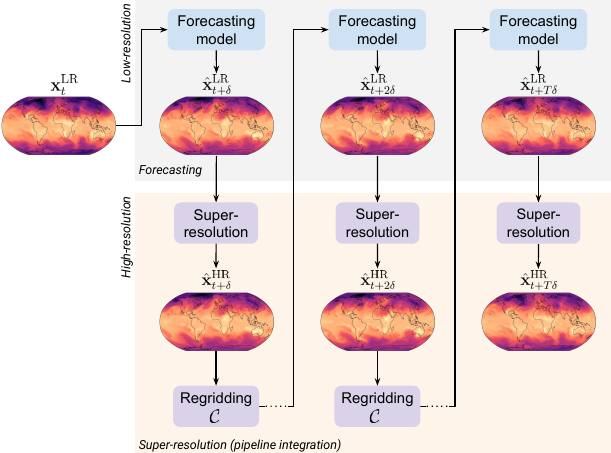}
      \caption{\textbf{Pipeline-integrated super-resolution.} Super-resolved fields are regridded and reinjected into the forecasting model, yielding an autoregressive coupling between super-resolution and forecasting.}
      \label{fig:sr_pipeline_integration}
\end{figure}

\emph{Pipeline integration.} A second mode of application consists in integrating super-resolution directly within the forecasting pipeline (Figure~\ref{fig:sr_pipeline_integration}). The motivation is to assess whether a more realistic representation of subgrid-scale variability, once re-aggregated to the coarse grid, may mitigate systematic biases and influence forecast evolution at longer lead times. In this setting, the super-resolved fields are re-coarsened and used to initialize subsequent forecast steps, yielding the autoregressive update
\begin{equation}
      \hat{\mathbf{x}}_{t+(k+1)\delta}^{\mathrm{LR}}
      =
      f_\theta\!\left(\mathcal{C}\!\left(\mathcal{C}^{-1}_\phi\left(\hat{\mathbf{x}}_{t+k\delta}^{\mathrm{LR}}\right)\right)\right).
\end{equation}

We compare the post-processing and pipeline-integration configurations using paired significance tests on daily verification metrics, following the statistical methods introduced in GenCast \citep{priceProbabilisticWeatherForecasting2025}.
For each variable and lead time, we compute paired time series of verification scores (CRPS, ensemble-mean RMSE, and Energy Score) across the 732 initialization dates of the 2020 evaluation period. Statistical significance is assessed using a paired stationary block bootstrap \citep{politisStationaryBootstrap1994} with automatic block-length selection \citep{politisAutomaticBlockLengthSelection2004, pattonCorrectionAutomaticBlockLength2009}, and confidence intervals are constructed using the bias-corrected and accelerated (BCa) method \citep{efronAutomaticConstructionBootstrap2020}. The null hypothesis of no difference between configurations is rejected when the 95\% confidence interval does not contain zero.

Across all conducted tests on all nine headline variables and lead times, statistically significant differences ($p < 0.05$) are observed primarily at intermediate lead times (2--5 days). This concentration likely reflects the autoregressive fine-tuning of ArchesWeatherGen over this lead-time range. When present, these differences tend to favor the post-processing configuration, but their magnitude remains small, with typical score differences on the order of $10^{-3}$.
Overall, these results indicate that reinjecting super-resolved states into the forecasting pipeline does not lead to systematic changes in global forecast skill.

\section{Zero-shot application to IFS HRES}
While all experiments in the main paper focus on super-resolving ML-based forecasts trained on ERA5, we expect the proposed super-resolution framework to be applicable, to some extent, to other forecasting systems. Numerical weather prediction outputs differ from ERA5 in their effective resolution, in particular through a different distribution of energy at small spatial scales. This setting provides an opportunity to examine the behavior of the super-resolution operator when applied outside its training distribution.

We apply the super-resolution model in a zero-shot manner to forecasts produced by IFS HRES \citep{81307}, without any retraining or fine-tuning. HRES forecasts at 0.25° resolution are first re-coarsened to 1.5° using the same first-order conservative regridding operator used throughout this study, and the resulting fields are provided as low-resolution inputs to the super-resolution model. To limit computational cost while retaining a representative sample of synoptic conditions, this experiment is conducted over a reduced test period in 2020, selecting initial times on the 1st and 15th day of each month at 00 and 12~UTC, for a total of 48 forecasts.

\begin{figure}[ht]
      \centering
      \includegraphics[width=\textwidth]{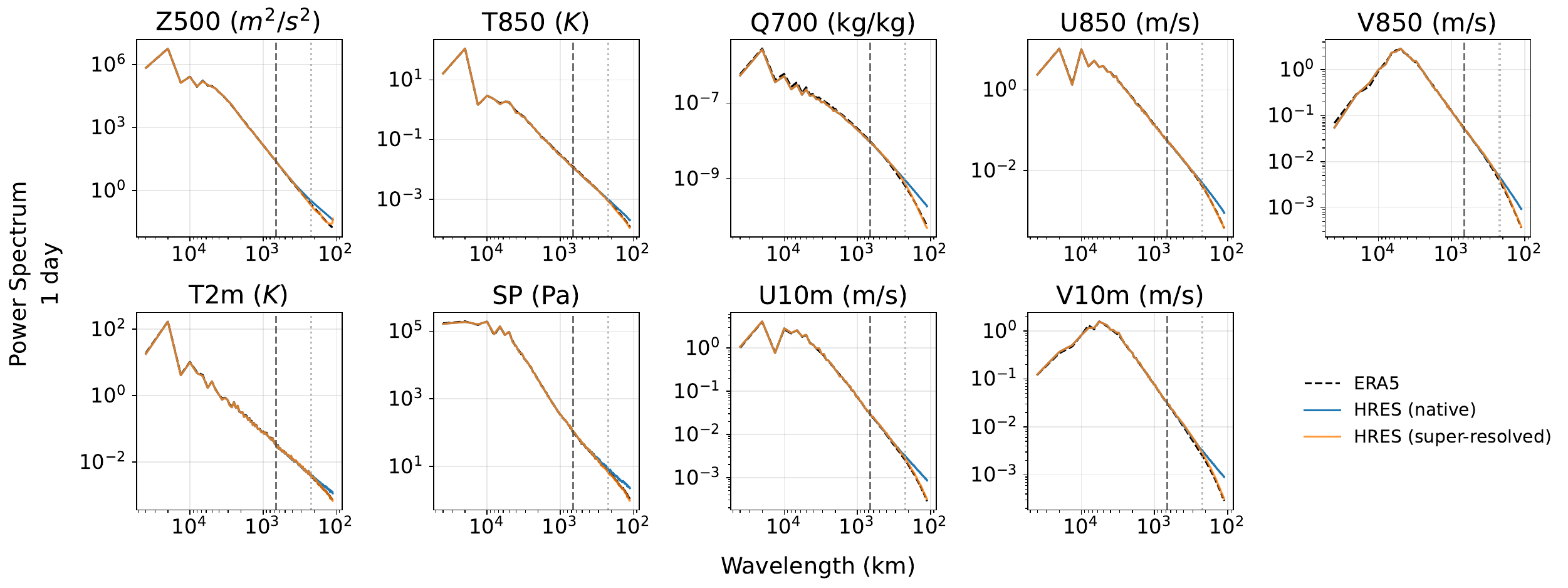}
      \caption{\textbf{Power spectra of IFS HRES at native and super-resolved resolutions at 1-day lead time.}
            Shown are spatial power spectra averaged over the test period for selected upper-air and surface variables. Spectra are displayed for ERA5, native IFS HRES forecasts, and IFS HRES fields obtained by applying the super-resolution model in a zero-shot setting. The vertical dashed line indicates the coarse-resolution cutoff at 1.5$^\circ$. The vertical dotted line indicates the change in effective resolution between native HRES and super-resolved HRES.}
      \label{fig:hres_spectrum_full_t+1}
\end{figure}

Figure~\ref{fig:hres_spectrum_full_t+1} shows spatial power spectra of ERA5, native IFS HRES forecasts, and the corresponding super-resolved fields at a 1-day lead time. At large spatial scales, the super-resolved fields closely follow the native IFS HRES spectra, indicating that the application of super-resolution does not modify the large-scale spectral content of the forecast. Differences between native and super-resolved IFS HRES fields emerge primarily at wavelengths shorter than approximately 220 km, where the super-resolved spectra depart from the native HRES behavior and approach the range of energy levels present in the ERA5 training data. From this perspective, the observed effect reflects the reconstruction of ERA5-like variability and may, depending on the intended use of the super-resolution model, act as a form of bias correction, although confirming this would require additional dedicated experiments. Targeting NWP-consistent small-scale variability would instead require training on high-resolution NWP analyses or reanalyses.

\end{document}